\documentclass[preprint,12pt,3p]{elsarticle}
\usepackage{amssymb}
\usepackage{booktabs} % For formal tables
\usepackage{amsmath,amssymb,amsfonts}
\usepackage[algoruled, lined, linesnumbered]{algorithm2e}
\usepackage{algcompatible}
\usepackage{caption}
\newsavebox\CBox
\def\textBF#1{\sbox\CBox{#1}\resizebox{\wd\CBox}{\ht\CBox}{\textbf{#1}}}
\usepackage{textcomp}
\usepackage{hyperref}
\usepackage{longtable}
\usepackage{multirow}
\usepackage{placeins}
\usepackage{multicol}
\usepackage{subfig}
\usepackage{wasysym}
\usepackage{enumitem}
\usepackage{cases}
\usepackage{cancel}
\usepackage[flushleft]{threeparttable}
\usepackage{setspace}

\usepackage{cuted}
\usepackage{color,soul}
\usepackage{soulutf8}
\usepackage{comment}

\usepackage{lineno}
\journal{Computers \& Operations Research}

\begin{document}

\begin{frontmatter}

\title{A weighted-sum method for solving the bi-objective traveling thief problem}
%\tnotetext[label0]{This is only an example}

\author[label1,label2]{Jonatas B. C. Chagas\corref{cor1}}
\ead{jonatas.chagas@iceb.ufop.br}

\author[label3]{Markus Wagner}
\ead{markus.wagner@adelaide.edu.au}

\address[label1]{Departamento de Computa\c{c}\~{a}o, Universidade Federal de Ouro Preto, Ouro Preto, Brazil}
\address[label2]{Departamento de Inform\'{a}tica, Universidade Federal de Vi\c{c}osa, Vi\c{c}osa, Brazil}
\address[label3]{School of Computer Science, The University of Adelaide, Adelaide, Australia}
\cortext[cor1]{Corresponding author.}
%\fntext[label3]{I also want to inform about\ldots}
%\fntext[label4]{Small city}
%\ead[url]{author-one-homepage.com}

\begin{abstract}
Many real-world optimization problems have multiple interacting components. 
Each of these can be $\mathcal{NP}$-hard and they can be in conflict with each other, i.e., the optimal solution for one component does not necessarily represent an optimal solution for the other components. 
This can be a challenge for single-objective formulations, where the respective influence that each component has on the overall solution quality can vary from instance to instance. 
In this paper, we study a bi-objective formulation of the traveling thief problem, which has as components the traveling salesperson problem and the knapsack problem. 
We present a weighted-sum method that makes use of randomized versions of existing heuristics, that outperforms participants on 6 of 9 instances of recent competitions, and that has found new best solutions to 379 single-objective problem instances.
\end{abstract}


\end{frontmatter}

%%\linenumbers

\sloppy

\section{Introduction}
\label{sec:introduction}

Real-world optimization problems often consist of several $\mathcal{NP}$-hard combinatorial optimization problems that interact with each other~\citep{klamroth2016interwoven,Bonyadi2019}. Such multi-component optimization problems are difficult to solve not only because of the contained hard optimization problems, but in particular, because of the interdependencies between the different components. Interdependence complicates decision-making by forcing each sub-problem to influence the quality and feasibility of solutions of the other sub-problems. This influence might be even stronger when one sub-problem changes the data used by another one through a solution construction process. Examples of multi-component problems are vehicle routing problems under loading constraints~\citep{iori2010routing,pollaris2015vehicle}, maximizing material utilization while respecting a production schedule~\citep{cheng2016supply,wang2020integrated}, and relocation of containers in a port while minimizing idle times of ships~\citep{forster2012tree,jin2015solving,hottung2020deep}.

In 2013, \cite{bonyadi2013travelling} introduced the traveling thief problem (TTP) as an academic multi-component problem. The academic `twist' of it is particularly important because it combines the classical traveling salesperson problem (TSP) and the knapsack problem (KP) -- both of which are very well studied in isolation -- and because of the interaction of both components can be adjusted. In brief, the TTP comprises a thief stealing items with weights and profits from a number of cities. The thief has to visit all cities once and collect items such that the overall profit is maximized. The thief uses a knapsack of limited capacity and pays rent for it proportional to the overall travel duration. To make the two components (TSP and KP) interdependent, the speed of the thief is made non-linearly dependent on the weight of the items picked so far.
The interactions of the TSP and the KP in the TTP result in a complex problem that is hard to solve by tackling the components separately.

The TTP has been gaining fast attention due to its challenging interconnected multi-components structure, and also propelled by several competitions\footnote{\url{https://cs.adelaide.edu.au/~optlog/research/combinatorial.php}} organized to solve it, which have led to significant progress in improving the performance of solvers. Among these, are iterative and local search heuristics~\citep{polyakovskiy2014comprehensive,faulkner2015approximate,maity2020efficient}, solution approaches based on co-evolutionary strategies~\citep{bonyadi2014socially,el2015cosolver2b,namazi2019cooperative}, memetic algorithms~\citep{mei2014interdependence,el2016population}, swarm-intelligence based approaches~\citep{wagner2016stealing,Zouari2019antstpp}, simulated annealing algorithm~\citep{el2018efficiently} and evolutionary strategy with probabilistic distribution model to construct high-valued solution from low-level heuristics~\citep{Martins2017ttpeda}. Exact approaches have also been considered, however they are limited to address very small instances~\citep{wu2017exact}.

As the TTP's components are interlinked, multi-objective considerations that investigate the interactions via the idea of ``trade-off''-relationships have been becoming increasingly popular. 
For example, \cite{yafrani2017ttpemo} created a fully-heuristic approach that generates diverse sets of solutions, while being competitive with the state-of-the-art single-objective algorithms. 
\cite{wu2018evolutionary} considered a bi-objective version of the TTP, which used dynamic programming as an optimal subsolver, where the objectives were the total weight and the TTP objective score. 
At two recent competitions\footnote{\textit{EMO-2019} \url{https://www.egr.msu.edu/coinlab/blankjul/emo19-thief/}}\textsuperscript{,}\footnote{\textit{GECCO-2019} \url{https://www.egr.msu.edu/coinlab/blankjul/gecco19-thief/}}, a bi-objective TTP (BITTP) variant has been used that trades off the total profit of the items and the travel time. 
The same BITTP variant was investigated by \cite{blank2017solvingBittp}, who proposed a simple algorithm for solving the problem. More recently, \cite{chagas2020nondominated} proposed a customized \mbox{NSGA-II} with biased random-key encoding. The authors have evaluated their algorithm on 9 instances, the same ones used in the aforementioned BITTP competitions. Their algorithm has shown to be effective according to the competition results.

In this work, we also address the BITTP variant used in competitions with a simple and effective heuristic approach. Specifically, our contributions with this paper are:

\setlist[itemize]{noitemsep, topsep=1pt}
\setlist[enumerate]{noitemsep, topsep=1pt}
\setitemize{noitemsep,topsep=4pt,parsep=0pt,partopsep=0pt}
\begin{enumerate}
    \item We have realized that we can decompose the multi-objective problem into a number of single-objective ones using a simple weighted-sum method~\citep{zadeh1963optimality}, which is one of the oldest strategies for addressing multi-objective optimization problems~\citep{ramanathan2006abc,marler2010weighted,stanimirovic2011linear,galand2012exact}.
    \item We tackle each single-objective problem through a two-stage heuristic by constructing a tour for the thief and then from it, we determine the packing plan with the stolen items. We use well-known efficient strategies for finding good tours and a problem-specific packing heuristic, which is a randomized version of a popular deterministic heuristic for the single-objective TTP, for determining the items stolen by the thief. 
    \item We incorporate into our algorithm the concepts of exploration and exploitation, which are aspects of effective search procedures~\citep{vcrepinvsek2013exploration,qi2015decomposition} by combining with our two-stages strategy, efficient local search operators already used in the single-objective TTP.
    \item To investigate the contributions that our algorithmic components have, we tune our solution method on 96 groups of instances and characterize the resulting configurations.
    \item We compare our approach with the tuned variant of \cite{chagas2020nondominated}, with the competition entries of the two competitions, and with single-objective TTP solvers.
\end{enumerate}

In the remainder of this article, we first define the BITTP in Section~\ref{sec:problem}. 
Then, in Section~\ref{sec:method}, we describe our weighted-sum method, where the decomposition is based on the respective influence of the two interacting components. There, we also introduce a randomized version of a popular packing strategy.
Section~\ref{sec:computational_experiments} contains the computational evaluation: the tuning of configurations and their characterisation, and the comparison with a range of (tuned) approaches from the literature, with the entries for two recent BITTP competitions, and with single-objective TTP solvers. Finally, in Section~\ref{sec:conclusions}, we present the conclusions and give suggestions for further investigations.

\section{Problem definition}\label{sec:problem}

The Bi-objective Traveling Thief Problem (BITTP) can be formally described as follows. There is a set of $m$ items $\{1, 2, \ldots, m\}$ distributed among a set of $n$ cities $\{1, 2, \ldots, n\}$. For any pair of cities $i, j \in \{1, 2, \ldots, n\}$, the distance $d(i,j)$ between them is known. Every city, except the first one, contains a subset of the $m$ items. Each item $j \in \{1, 2, \ldots, m\}$ has a profit $p_j$ and a weight $w_j$ associated. There is a single thief who has to visit all cities exactly once starting from the first city and returning back to it in the end (TSP component). The thief can make a profit by stealing items and storing them in a knapsack with a limited capacity $W$ (KP component). As stolen items are stored in the knapsack, it becomes heavier, and the thief travels more slowly, with a velocity inversely proportional to the knapsack weight. Specifically, the thief can move with a speed $v = v_{max} - w \times (v_{max} - v_{min})\,/\, W$, where $w$ is the current weight of their knapsack. Consequently, when the knapsack is empty, the thief can move with the maximum speed $v_{max}$; when the knapsack is full, the thief moves with the minimum speed $v_{min}$.

Any feasible solution for the BITTP can be represented through a pair $\langle\pi, z\rangle$, where $\pi = \langle\pi_1, \pi_2, \ldots, \pi_n\rangle$ is a permutation of all cities in the order they are visited by the thief, and $z = \langle z_1, z_2, \ldots, z_m \rangle$ is a binary vector representing the packing plan ($z_j = 1$ if item $j$ is collected, and $0$ otherwise) adopted by the thief throughout their robbery journey. We can formally express the space of feasible solutions for the BITTP by constraints \eqref{eq:do_not_repeat_cities} to \eqref{eq:pick_or_not_each_item}.

\begin{align}
    \label{eq:do_not_repeat_cities}
    &\pi_i \neq \pi_j &i \in \{1, 2, \ldots, n\},j \in \{i+1, i+2, \ldots, n\} \\
    \label{eq:candidate_cities}
    &\pi_{i} \in \{1, 2, \ldots, n\} &i \in \{1, 2, \ldots, n\} \\
    \label{eq:first_city}
    &\pi_1 = 1 \\
    \label{eq:knapsack_constraint}
    &\sum_{j=1}^m z_j \cdot w_j \leq W \\
    \label{eq:pick_or_not_each_item}
    &z_{j} \in \{0, 1\} &j \in \{1, 2, \ldots, m\}
\end{align}

Constraints~\eqref{eq:do_not_repeat_cities} and \eqref{eq:candidate_cities} ensure that each city is visited exactly once, while constraint \eqref{eq:first_city} establishes that the thief must start their journey from city 1. Constraints~\eqref{eq:knapsack_constraint} and \eqref{eq:pick_or_not_each_item} ensure, respectively, that the knapsack capacity is not exceeded, and that each item may be collected only once.

The objectives of the BITTP are to maximize the profit of the collected items and to minimize the total traveling time spent by the thief to conclude their journey. These objectives are mathematically defined according to Equations~\eqref{eq:obj_profit} and \eqref{eq:obj_time}, respectively.

\begin{align}
    \label{eq:obj_profit}
    &\max\;g(z)\;=\;\sum_{j=1}^{m}  p_j \cdot z_j \\
    \label{eq:obj_time}
    &\min\;h(\pi, z)\;=\;\sum_{i=1}^{n-1} \frac{d(\pi_i, \pi_{i+1})}{v_{max} - v \cdot \omega(i,\pi,z)}  \; + \; \; \frac{ d(\pi_n, \pi_{1})}{v_{max} - v \cdot \omega(n,\pi,z)}
\end{align}

Note that while the objective \eqref{eq:obj_profit} is calculated directly from the packing plan $z$, the calculation of the objective \eqref{eq:obj_time} is more complex. Since the speed of the thief depends on the current weight of their knapsack, it may change after visiting each city. Therefore, it is necessary to know the traveling speed between each pair of cities in order to calculate the total traveling time. For this purpose, it is necessary to determine the total weight of the knapsack after visiting each city $i$ according to the tour $\pi$ and the packing plan $z$, which is denoted by $\omega(i,\pi,z)$ and is calculated as described in Equation~\eqref{eq:total_weight}. Hence, all speeds of the thief throughout their journey, and, consequently, the total traveling time can be computed.
\begin{equation}
    \label{eq:total_weight}
    \omega(i,\pi,z) = \sum_{k=1}^{i} \sum_{j=1}^{m} w_j \cdot z_j \cdot  \begin{cases}
   \;1 & \text{if item } j \text{ is localized in city } \pi_k \\
   \;0 & \text{otherwise}
  \end{cases}
\end{equation}

It is important to note that the objectives of the BITTP are conflicting with each other, as optimizing each one of them independently does not necessarily produce a good solution in terms of the other objective. Indeed, for faster tours, the thief should not collect items or collect a few items with small weights. On the other hand, for collecting sets of items with high profit, the thief travels slowly due to the weight of the collected items. Therefore, there is no single solution that simultaneously optimizes both objectives, but a set of solutions, called Pareto-optimal solutions, in which each solution is non-dominated in terms of its objective values by any other solution.

\section{Problem-solving methodology}\label{sec:method}

Throughout this section, we discuss the methodology we have adopted in order to find high-quality non-dominated solutions for the BITTP. We describe in detail all components of our proposed algorithm as well as all the decisions made during its design development.

\subsection{The overall algorithm}

Our proposed algorithm is based on the weighted-sum method (WSM)~\citep{zadeh1963optimality}, a well-known strategy for addressing multi-objective optimization problems~\citep{marler2010weighted}. Basically, its core idea consists of converting the multi-objective problem at hand into several single-objective problems by using different convex combinations of the original objectives. Then, each one of the created single-objective problems is solved in order to generate non-dominated solutions for the multi-objective problem~\citep{das1997closer}. Note that the optimal solution for each single-objective problem is a Pareto-optimal solution for the multi-objective problem, because, if this were not the case, then there must exist another feasible solution with an improvement on at least one of the objectives without worsening the others. Hence, that solution would have a better value according to the weighted-sum objective function.

According to \cite{marler2010weighted}, the WSM is often used for addressing real-world applications, especially for those with just two objective functions, not only to provide multiple solutions widely spread across the space of the objectives, but also to provide a single solution that reflects preferences presumably incorporated in the selection of a single set of weights for the objectives. WSM has also given rise to very popular multi-objective decomposition-based optimization algorithms like MOEA/D~\citep{Zhang2007moead}.

Limitations of WSMs include their inability to capture Pareto-optimal solutions that lie in non-convex portions of the Pareto-optimal curve, and also that they do not necessarily generate a dispersed distribution of solutions in the Pareto-optimal set, even with a consistent change in weights attributed to the objectives. Throughout the article, we point out why these limitations do not affect our algorithm.

For the BITTP, our proposed WSM converts the objective functions \eqref{eq:obj_profit} and \eqref{eq:obj_time} into the weighted-sum objective function \eqref{eq:weighted_sum_obj} by including a scalar value $\alpha$ that may assume any real number between 0 and 1. In addition, we have included in weighted-sum objective function the renting rate $R$ defined by \cite{polyakovskiy2014comprehensive} for the set of TTP instances, which is widely used as benchmarking in TTP related researches. As stated by \cite{polyakovskiy2014comprehensive} the renting value has been tailored to each TTP instance, and its value establishes the connection between both TTP components. It is important to emphasize that the renting values vary widely among the benchmarking TTP instances. Thus, by varying the $\alpha$ values, we will be creating new TTP instances with different weights/importance for their components, but they will still have the tailored influence of the renting rate. 
\begin{equation}
    \label{eq:weighted_sum_obj}
    \max \;\; f(\pi, z, \alpha) =  \alpha \cdot g(z) - (1 - \alpha) \cdot R \cdot h(\pi, z)
\end{equation}

Although exact algorithms exist for the TTP, they are limited to solving very small instances within a reasonable computational time \citep{wu2017exact}. In fact, unless $\mathcal{P} = \mathcal{NP}$, it is not possible to develop an exact strategy able to solve general TTP instances in polynomial time. Therefore, we solve each new TTP instance by using concepts of effective heuristic approaches proposed for the TTP over the years. Consequently, there is no guarantee that our WSM finds Pareto-optimal solutions. On the other hand, it is able to find solutions possibly located in non-convex portions of the Pareto-optimal curve. Indeed, there is no convex combination of the two objectives whose global optimal value corresponds a solution located in non-convex portions. However, since each single-objective problem is approached with a heuristic strategy, these solutions can be achieved when the heuristic fails to find the global optimal value.

As the TTP has gained increasing attention since its proposition, several approaches have emerged to solve it. Some of them use techniques that require higher computational effort, whereas others bet on low-level search operators, which can also produce high-quality solutions with shorter computation time \citep{polyakovskiy2014comprehensive, faulkner2015approximate, wagner2018case}. As the BITTP demands a set of non-dominated solutions instead of a single solution, a higher computational effort is required to find high-quality solutions. Thus, we have designed our solution strategy with low-level search operators in mind with the purpose of develop an efficient and scalable solution approach that balances the concepts of exploration and exploitation in order to find high-quality and high-diversity non-dominated BITTP solutions. 

In Algorithm~\ref{alg:wsm}, we present in detail the steps performed in our WSM for solving the BITTP. It starts (Line \ref{alg:wsm_init_nd_set}) by initializing the set that stores all non-dominated solutions found throughout the algorithm. By non-dominated solutions, we refer to solutions $\mathcal{S} \subseteq \mathcal{S}'$, from the set of solutions $\mathcal{S}'$ that our algorithm found, where none of the solutions from $\mathcal{S}' \setminus \{\mathcal{S}\}$ dominates the solutions from $\mathcal{S}$. Our algorithm performs iterative cycles (Lines \ref{alg:wsm_loop_begin} to \ref{alg:wsm_loop_end}) while its stopping criterion is not achieved. At each iteration, we carry out exploration and exploitation mechanisms. During the exploration phase (Lines \ref{alg:wsm_exploration_begin} to \ref{alg:wsm_exploration_end}), our algorithm generates $\eta$ feasible solutions for the BITTP as follows. Initially (Line \ref{alg:wsm_clhk}), a tour $\pi$ is generated by using the well-known Chained-Lin-Kernighan heuristic \citep{applegate2003chained}. 

Afterwards, we construct a feasible packing plan $z$ at a time (Line~\ref{alg:wsm_randomized_packing_algorithm}) by using a randomized packing heuristic we have developed. Then, each packing plan $z$ is combined with tour $\pi$ in order to compose a feasible solution $\langle\pi,z\rangle$ for the BITTP, which is used to update the set of non-dominated solutions $\mathcal{S}$ (Line~\ref{alg:wsm_udpate_nds}). The update of $\mathcal{S}$ is done in order to keep only non-dominated solutions in the set. Thus, if a solution $\langle\pi,z\rangle$ is dominated by any solution in $\mathcal{S}$, it is discarded. Otherwise, $\langle\pi,z\rangle$ is added in $\mathcal{S}$ and all solutions dominated by it are then removed. All the details of our packing heuristic strategy will be presented later in Algorithm~\ref{alg:randomized_packing_algorithm}. For now, we would like to only stress that each packing plan is constructed based upon the tour $\pi$ and also on the real number $\alpha$ used to define the current weighted-sum objective function. Note that, in our algorithm, a value for $\alpha$ is randomly generated from a probability distribution $\mathcal{D}$ (Line~\ref{alg:wsm_generate_alpha}). Thus, we can control and emphasize in which intervals of values $\alpha$ should be chosen by using different probability distributions. 

The exploitation phase (Lines \ref{alg:wsm_exploitation_begin} to \ref{alg:wsm_exploitation_end}) begins by generating a new $\alpha$ value (Line~\ref{alg:wsm_generate_alpha_2}) and selecting the best non-dominated solution $\langle\pi',z'\rangle$ in $\mathcal{S}$ according to the weighted-sum objective function formed from this new $\alpha$. The solution $\langle\pi',z'\rangle$ is considered as a pivot for applying two local operators: $2$-opt and bit-flip. Basically, a $2$-opt move removes two non-adjacent edges and inserts two new edges by inverting two parts of the tour in such a way that a new tour is formed. In turn, a bit-flip move inverts the state of an item $j$ in the packing plan $z'$, i.e., if $j$ is in $z'$ then it is removed; otherwise, it is inserted if its inclusion does not exceed the knapsack capacity. These operators have been successfully incorporated to solve various combinatorial optimization problems, including the single-objective TTP~\citep{faulkner2015approximate, el2016population, chand2016fast, el2018efficiently}, and also the BITTP~\citep{chagas2020nondominated}. 

In our algorithm, first, we apply the operator $2$-opt over the tour $\pi'$ while the packing plan $z'$ remains unchanged in order to find a faster tour that is still able to collect the same set of items. As the number of all tours $\Pi$ (Line~\ref{alg:wsm_set_all_tours}) obtained from $2$-opt moves may be huge for some instances, it is impracticable to analyze them all. In addition, significantly longer tours have less potential to be faster. For that reason, our algorithm has been restricted to analyze only those tours that are longer than $\pi'$ up to a limited distance (Lines~\ref{alg:wsm_select_shorter_tours_begin} to \ref{alg:wsm_select_shorter_tours_end}). The maximum tolerance for accepting a tour is given by the average of the distance $\ell$ among all pair of cities multiplied by a factor $\beta$ (Line~\ref{alg:wsm_select_shorter_tours_if}). After analyzing all selected tours, we chose the fastest tour $\pi''$, if any, among those that are faster than $\pi'$, to compose a new solution, and then the set of non-dominated solutions $\mathcal{S}$ is updated from it (Line~\ref{alg:wsm_update_nds_with_2opt_tour}). 

Afterwards, bit-flip operations are applied to the packing plan $z'$ in order to find new packing plans that when combined with the tour $\pi'$ produce new solutions. Because generating all bit-flip moves and evaluating all solutions formed from them may be impracticable for instances with many items, we decided that each bit-flip move is done according to a probability $\lambda$ (Lines~\ref{alg:wsm_bitflip_moves_begin} to \ref{alg:wsm_bitflip_moves_end}). The solutions generated from bit-flip moves are used to update, if applicable, the set of non-dominated solutions $\mathcal{S}$ (Line~\ref{alg:wsm_update_nds_with_bitflip_packing_plans}). At the end of the algorithm (Line~\ref{alg:wsm_return}), all non-dominated solutions found throughout its execution are returned.

\begin{algorithm}%[t]
\setstretch{1.25}
\makeatletter
\newcommand{\algorithmfootnote}[2][\footnotesize]{%
  \let\old@algocf@finish\@algocf@finish% Store algorithm finish macro
  \def\@algocf@finish{\old@algocf@finish% Update finish macro to insert "footnote"
    \leavevmode\rlap{\begin{minipage}{\linewidth}
    #1#2
    \end{minipage}}%
  }%
}
%\tiny
%\scriptsize
\footnotesize
%\small
%\normalsize
\DontPrintSemicolon
\SetKwData{Left}{left}
\SetKwData{Up}{up}
\SetKwFunction{FindCompress}{FindCompress}
\SetKwInOut{Input}{input}
\SetKwInOut{Output}{output}
%\Input{A bitmap $Im$ of size $w\times l$}
%\Output{A partition of the bitmap}
%\BlankLine
$\mathcal{S} \gets \varnothing$ \quad \tcp{set of non-dominated solutions} \label{alg:wsm_init_nd_set}
\Repeat{\upshape stopping condition is fulfilled} { \label{alg:wsm_loop_begin}
    \tcp{exploration phase:} 
    $\pi \gets$ solve the TSP component by the Chained-Lin-Kernighan  heuristic$^{\spadesuit}$ \label{alg:wsm_exploration_begin} \label{alg:wsm_clhk} \\
    \For{\upshape $k \gets 1 \textbf{ to } \eta$} {
        $\alpha \gets$ generate a random number from the probability distribution $\mathcal{D}$ \label{alg:wsm_generate_alpha} \\
        $z \gets \textsc{RandomizedPackingAlgorithm}(\pi,\,\rho,\,\alpha,\,\gamma)$ \tcp*{Algorithm \ref{alg:randomized_packing_algorithm}} \label{alg:wsm_randomized_packing_algorithm}
        update $\mathcal{S}$ with the solution $\langle\pi, z\rangle$ \label{alg:wsm_udpate_nds}
    } \label{alg:wsm_exploration_end}
    \tcp{exploitation phase:}
    $\alpha \gets$ generate a random number from the probability distribution $\mathcal{D}$ \label{alg:wsm_exploitation_begin} \label{alg:wsm_generate_alpha_2} \\
    $\langle\pi', z'\rangle \gets $ get from $\mathcal{S}$ the best solution according to $\alpha$ \\
    \tcp{exploitation phase (2-opt moves):}
    let $\Pi$ be the set of all 2-opt tours obtained from $\pi'$ \label{alg:wsm_2opt_moves_begin} \label{alg:wsm_set_all_tours} \\
    let $\ell$ be the average of the distance among all pair of cities \\
    $\pi'' \gets \pi'$ \\
    \ForEach{$\pi''' \in \Pi$} { \label{alg:wsm_select_shorter_tours_begin} 
        \If{$d(\pi''') - d(\pi') \leq \ell \times \beta$} { \label{alg:wsm_select_shorter_tours_if}
            \lIf{$f(\pi''', z', \alpha) > f(\pi'', z', \alpha)$} {
                $\pi'' \gets \pi'''$
            }
        }
    } \label{alg:wsm_select_shorter_tours_end}
    \lIf{$\pi'' \neq \pi'$} {
        update $\mathcal{S}$ with the solution $\langle\pi'', z'\rangle$ \label{alg:wsm_update_nds_with_2opt_tour}
    } \label{alg:wsm_2opt_moves_end}
    \tcp{exploitation phase (bit-flip moves):}
    \ForEach{\upshape item $j \in \{1, 2, \ldots, m\}$} { \label{alg:wsm_bitflip_moves_begin}
        \If{$rand(0, 1) \leq \lambda$} {
            \If{$j \in z'$} { 
                update $\mathcal{S}$ with the solution $\langle\pi', z' \setminus \{j\}\rangle$
            } 
            \ElseIf{\upshape weight of $z' \cup \{j\}$ is lower than $W$} {
                update $\mathcal{S}$ with the solution $\langle\pi', z' \cup \{j\}\rangle$ \label{alg:wsm_update_nds_with_bitflip_packing_plans}
            }
        }
    } \label{alg:wsm_exploitation_end} \label{alg:wsm_bitflip_moves_end}
} \label{alg:wsm_loop_end}
\Return $\mathcal{S}$ \label{alg:wsm_return}
\caption{Weighted-Sum Method -  \textsc{WSM}($\mathcal{D},\,\eta,\,\rho,\,\gamma,\,\beta,\,\lambda$)}
\label{alg:wsm}
\algorithmfootnote{\scriptsize $\spadesuit$ \url{http://www.math.uwaterloo.ca/tsp/concorde/downloads/downloads.htm}}
\end{algorithm}%

\subsection{A randomized packing strategy}

In order to complete the description of the proposed WSM, we now present the strategy used to generate a packing plan from a given tour $\pi$. It is important to highlight that even for this scenario, the task of finding the optimal packing configuration remains $\mathcal{NP}$-hard~\citep{polyakovskiy2015packing}, which makes it impractical for any but the smallest instances\footnote{at least with the methods known to date~\citep{wu2017exact}} due to the time of computing required, especially because this procedure is a subroutine of our entire algorithm that is called many times. For this reason, our proposed strategy is a heuristic approach with the aim of quickly obtaining a packing plan from a tour. Before presenting its details, we would like to emphasize that our strategy is a non-deterministic packing algorithm, i.e., even for the same input parameters, it may exhibit different behaviors on different runs. Our design decision for that has been based on the fact that a non-deterministic mechanism introduces a more broadly exploration of the packing plan space, which may be effective to find regions with high-quality solutions.

\begin{algorithm}[!ht]
\makeatletter
\newcommand{\algorithmfootnote}[2][\footnotesize]{%
  \let\old@algocf@finish\@algocf@finish% Store algorithm finish macro
  \def\@algocf@finish{\old@algocf@finish% Update finish macro to insert "footnote"
    \leavevmode\rlap{\begin{minipage}{\linewidth}
    #1#2
    \end{minipage}}%
  }%
}
%\tiny
%\scriptsize
\footnotesize
%\small
%\normalsize
\DontPrintSemicolon
\SetKwData{Left}{left}
\SetKwData{Up}{up}
\SetKwFunction{FindCompress}{FindCompress}
\SetKwInOut{Input}{input}
\SetKwInOut{Output}{output}
%\Input{A bitmap $Im$ of size $w\times l$}
%\Output{A partition of the bitmap}
%\BlankLine
$z^{best} \gets \varnothing$ \label{alg:random_packing_init_z_best} \\
\For{\upshape $\rho' \gets 1 \textbf{ to } \rho$} { \label{alg:random_packing_for_rho_begin}
    $a \gets rand(0, 1)$, $b \gets rand(0, 1)$, $c \gets rand(0, 1)$  \label{alg:random_packing_random_a_b_c_values} \label{alg:random_packing_attempt_begin} \\
    normalize $a$, $b$, and $c$ so that their sum is equal to 1 \label{alg:random_packing_normalize_a_b_c_values} \\
    compute score for each item using $a$, $b$, and $c$ according to Eq. \eqref{eq:item_score} \label{alg:random_packing_item_score} \\
    $\varphi \gets \lceil m / \gamma \cdot \alpha + \epsilon\rceil$ \label{alg:random_packing_init_varphi} \\
    $z \gets z' \gets \varnothing$ \label{alg:random_packing_init_z_zline} \\
    $\text{newPackingPlan} \gets \textbf{false}$ \label{alg:random_packing_var_new_packing_plan} \\
    $k \gets k' \gets 1$ \label{alg:random_packing_var_k_kline}  \\
    \While{\upshape $k' \leq m$ \textbf{and} $\varphi \geq 1$} { \label{alg:random_packing_iterative_packing_construction_begin}
        $j \gets $ get item with the $k'$-th largest score \label{alg:random_packing_select_item_kline} \\
        \If{\upshape weight of $z' \cup \{j\}$ is lower than $W$} { \label{alg:random_packing_if_collect_item}
            $z' \gets z' \cup \{j\}$, $\text{newPackingPlan} \gets \textbf{true}$ \label{alg:random_packing_collect_item}
        }
        \If{\upshape $k'$ \textbf{mod} $\varphi = 0$ \textbf{and} $\text{newPackingPlan} = \textbf{true}$} { \label{alg:random_packing_if_new_packing}
            \If{$f(\pi, z', \alpha) > f(\pi, z, \alpha) $} { \label{alg:random_packing_zline_is_better}
                $z \gets z'$,\, $k \gets k'$ 
            }
            \lElse { \label{alg:random_packing_zline_is_not_better}
                $z' \gets z$,\, $k' \gets k,\, \varphi \gets \lfloor \varphi / 2\rfloor$  \label{alg:random_packing_varphi_halved} 
            }
            $\text{newPackingPlan} \gets \textbf{false}$
        }
        $k' \gets k' + 1$
    } \label{alg:random_packing_attempt_end} \label{alg:random_packing_iterative_packing_construction_end}
    \lIf{\upshape $f(\pi, z, \alpha) > f(\pi, z^{best}, \alpha) $} { \label{alg:random_packing_attempt_is_better}
        $z^{best} \gets z$ 
    }
} \label{alg:random_packing_for_rho_end}
\Return $z^{best}$ \label{alg:random_packing_return}
\caption{\textsc{RandomizedPackingAlgorithm}($\pi,\,\rho,\,\alpha,\,\gamma$)}
\label{alg:randomized_packing_algorithm}
\end{algorithm}%

Algorithm~\ref{alg:randomized_packing_algorithm} describes all the steps of our packing heuristic strategy. It seeks to find a good packing plan $z^{best}$ from multiple attempts for the same tour $\pi$. At each attempt (Line~\ref{alg:random_packing_attempt_begin} to \ref{alg:random_packing_attempt_end}), a packing plan $z$ is constructed. Due to the non-deterministic nature of our packing algorithm, multiple attempts increase the chance of finding a better packing plan. The number of attempts can be controlled by the parameter $\rho$ (Line~\ref{alg:random_packing_for_rho_begin}). Before any of these attempts (Line~\ref{alg:random_packing_init_z_best}), $z^{best}$ is defined with no items. Afterwards, at the beginning of each attempt, we uniformly select three random values ($a$, $b$, and $c$) between 0 and 1 (Line~\ref{alg:random_packing_random_a_b_c_values}), and then normalize them (Line~\ref{alg:random_packing_normalize_a_b_c_values}) so that their sum is equal to 1. These values are used to compute a score $s_j$ for each item $j \in \{1, \ldots, m\}$ (Line~\ref{alg:random_packing_item_score}), where $a$, $b$, and $c$ define, respectively, exponents applied to profit $p_{j}$, weight $w_j$, and distance $d_{j}$ in order to manage their impact. The distance $d_{j}$ is calculated according to the tour $\pi$ by summing all the distances from the city where item $j$ is located to the final city of the tour. Equation~\ref{eq:item_score} shows how the score of item $j$ is calculated:
\begin{equation}
    \label{eq:item_score}
    s_{j} = \frac{(p_{j})^{a}}{(w_{j})^{b} \cdot (d_{j})^{c}}
\end{equation}

From the foregoing equation, we can note that each score $s_j$ incorporates a trade-off among a distance that item $j$ has to be carried over, its weight, and also its profit. Equation \ref{eq:item_score} is based on the heuristic \textsc{PackIterative} that has been developed for the TTP \citep{faulkner2015approximate}. However, unlike these last authors, we have also considered an exponent for the term of distance to vary the importance of its influence. Furthermore, the values of all exponents are randomly selected drawn between 0 and 1, and then they are normalized in such a way that each of them establishes a percentage of importance in the calculation of the score. After computing all scores, our algorithm uses their values to define the priority of each item in the packing strategy. The higher the score of an item, the higher its priority. 

As described in the following, each packing plan $z$ is constructed by selecting items iteratively according to their priorities. After including any item in $z$, it would be necessary to calculate the objective value of the solution $\langle\pi, z\rangle$ to be sure about its quality. However, since evaluating the objective function many times may be time-consuming, especially for large-size instances, we have introduced a parameter $\varphi$ for controlling the frequency of the objective value re-computation. In other words, the objective value of the current solution $\langle\pi, z\rangle$ is only evaluated each time that $\varphi$ items are analyzed. Initially (Line~\ref{alg:random_packing_init_varphi}), $\varphi$ is defined as $\lceil m / \gamma \cdot \alpha + \epsilon\rceil$, which depends on the number of items $m$, a parameter $\gamma$ and the value $\alpha$, and also a small value $\epsilon = 10^{-5}$ to avoid that $\varphi$ assumes 0 when $\alpha$ is 0. Thus, the lower $\alpha$, the lower $\varphi$ and, consequently, the higher the frequency of the objective value re-computation. Note that for values close to zero, we look for solutions with faster tours, which requires a packing plan without or with few items. Therefore, for this scenario, a high frequency of re-computation of the objective function is needed in order to select many items without checking whether they improve the quality of the solution.

Each packing plan $z$ is constructed as follows. At first, $z$ and an auxiliary packing plan $z'$ are both defined as empty sets (Line~\ref{alg:random_packing_init_z_zline}). Other auxiliary variables are used to control if there is a new packing plan to be evaluated (Line~\ref{alg:random_packing_var_new_packing_plan}) and also to management which item is currently being analyzed (Line~\ref{alg:random_packing_var_k_kline}). The iterative packing construction process of our algorithm (Lines~\ref{alg:random_packing_iterative_packing_construction_begin} to \ref{alg:random_packing_iterative_packing_construction_end}) start by selecting the item $j$ with the $k'$-th largest score (Line~\ref{alg:random_packing_select_item_kline}). If the addition of item $j$ does not exceed the knapsack capacity (Line~\ref{alg:random_packing_if_collect_item}), then $j$ is inserted into packing plan $z'$, and it is marked that there is a new packing configuration (Line~\ref{alg:random_packing_collect_item}). Every time that $\varphi$ items have been considered and that the current packing plan $z'$ has not been evaluated (Line~\ref{alg:random_packing_if_new_packing}), we compute the objective function of the solution $\langle\pi, z'\rangle$ and confront its quality against quality of the solution $\langle\pi, z\rangle$. If the solution $\langle\pi, z'\rangle$ is better (Line~\ref{alg:random_packing_zline_is_better}), $\varphi$ remains the same and $z$ is updated to $z'$. Otherwise (Line~\ref{alg:random_packing_zline_is_not_better}), the packing plan $z'$ is updated to $z$ and the algorithm returns to consider the items again starting with the item whose score is the $k$-th largest (Line~\ref{alg:random_packing_varphi_halved}). In addition, $\varphi$ is halved in order to provide the chance to improve the solution by collecting fewer items before an evaluation. Each construction of a packing plan terminates either when there is no more items to collect or because no further improvement is possible following our strategy. After completing the construction of each packing plan $z$, the best solution $\langle\pi, z^{best}\rangle$ found so far is updated to the solution $\langle\pi, z\rangle$ if it is to improve (Line~\ref{alg:random_packing_attempt_is_better}). At the end of the algorithm (Line~\ref{alg:random_packing_return}), the packing plan of the best solution found is returned.

\section{Computational experiments}
\label{sec:computational_experiments}

In this section, we present the experiments performed to study the performance of the proposed algorithm. First, we have conducted an extensive comparison with the algorithm proposed by \cite{chagas2020nondominated}. In addition, we compare our results with those submitted to BITTP competitions, which have been held in 2019 at the \emph{Evolutionary Multi-Criterion Optimization} (EMO2019) and \emph{The Genetic and Evolutionary Computation Conference} (GECCO2019). Lastly, we contrast our results with the single TTP objective scores obtained from efficient algorithms already proposed in the literature for the TTP.

Our algorithm has been implemented in Java. Each run of it has been sequentially (nonparallel) performed on a machine with Intel(R) Xeon(R) CPU X5650 @ 2.67GHz and Java 8, running under CentOS 7.4. Our code, as well as all numerical results, can be found at 
\href{https://github.com/jonatasbcchagas/wsm_bittp}{\textcolor{blue}{https://github.com/jonatasbcchagas/wsm\_bittp}}.

\subsection{Benchmarking instances}
\label{sec:ttp_instances}

To assess the quality of the proposed WSM, we have used instances of the comprehensive set of TTP instances defined by \cite{polyakovskiy2014comprehensive}. These authors have created 9720 instances in such a way that the two components of the problem have been balanced so that the near-optimal solution of one sub-problem does not dominate over the optimal solution of another sub-problem. For a complete and detailed description of how these instances have been created, we refer the interested reader to~\citep{polyakovskiy2014comprehensive} and also to~\citep{wagner2018case}, which presents a study on the instance features. In our experiments, we have used a subset of the 9720 TTP instances with the following characteristics:

\begin{itemize}
    \item {
        numbers of cities: 51, 152, 280, 1000, 4461, 13509, 33810, and 85900 (the layout of cities is given according to the TSP instances~\cite{reinelt1991tsplib} {\it eil51}, {\it pr152}, {\it a280}, {\it dsj1000}, {\it usa13509}, {\it pla33810}, and {\it pla85900}, respectively);
    }
    \item {
        numbers of items per city: {\it 01}, {\it 03}, {\it 05}, and {\it 10} (all cities of a single TTP instance have the same number of items, except for the city in which the thief starts and ends their journey, where no items are available);
    }
    \item {
        types of knapsacks: weights and values of the items are bounded and strongly correlated ({\it bsc}), uncorrelated with similar weights ({\it usw}), uncorrelated ({\it unc});
    }
    \item {
        sizes of knapsacks: {\it 01}, {\it 02}, {\it $\ldots$}, {\it 09} and {\it 10} times the size of the smallest knapsack, which is defined by summing the weight of all items and dividing the sum by 11, as per \cite{polyakovskiy2014comprehensive};
    }
\end{itemize}

By combining all the different characteristics described above, we have 960 instances that compose a broad and diverse sample of all 9720 instances. In the remainder of this article, each instance will be identified as {\tt XXX\_YY\_ZZZ\_WW}, where {\tt XXX}, {\tt YY}, {\tt ZZZ}, and {\tt WW} indicate the different characteristics of the instance at hand. For example,  {\tt a280\_03\_bsc\_01} identifies the instance with 280 cities (TSP instance {\it a280}), 3 items per city with their weights and values bounded and strongly correlated with each other, and the smallest knapsack defined.

\subsection{Parameter tuning}
\label{sec:parameter_tuning}

In order to find suitable configuration values for the algorithm's parameters among all possible ones, we have used the Irace package~\citep{lopez2016irace}, which is an implementation of the method I/F-Race~\citep{birattari2010f}. The Irace package implements an iterated racing framework for the automatic configuration of algorithms, which has been used frequently due to its simplicity to use and its performance.

Table~\ref{table:parameter_values} shows the parameter values of our algorithm we have considered in the Irace tuning. These values have been selected following preliminary experiments. Note that for $\beta = -\infty$ and $\lambda = 0$, our algorithm does not perform, respectively, any 2-opt and bit-flip moves. Regarding the stopping criterion of the algorithm, we have set its runtime to 10 minutes. This choice is very often used in TTP research, thus following a pattern already established that allows fairer comparisons among different solution approaches. In addition, as stated by \cite{wagner2018case}, this computation budget limit is motivated by a real-world scenario, where a 10-minutes break is enough for a decision-maker, who is interested in what-if analyses, to have a cup of coffee. After this time, the decision-maker analyses the computed results, and then he/she can make the possible next changes to the system to investigate other alternatives.

\begin{table}[!ht]
\centering
%\tiny
%\scriptsize
\footnotesize
%\small
%\normalsize
\centering
\caption{Parameter values considered during the tuning experiments.}
\setlength{\tabcolsep}{0pt}
\begin{tabular*}{\hsize}{@{}@{\extracolsep{\fill}}cl@{}}
\toprule
\textbf{Parameter} & \textbf{Tested values}\\ 
\midrule
$\mathcal{D}$ & $\mathcal{U}(0,1)$,\; $\mathcal{N}(0.5, 0.2)$,\; $\mathcal{B}(3, 1.5)$,\; $\mathcal{B}(1.5, 3)$  \\[1mm]
$\eta$ & $1, 2, \ldots, 200$ \\[1mm]
$\rho$ & $1, 2, \ldots, 100$ \\[1mm]
$\gamma$ & $1, 2, \ldots, 200$ \\[1mm]
$\beta$ & $-\infty, 0, 0.000001, 0.00001, 0.0001, 0.001, 0.01, 0.1, 10, 100$ \\[1mm]
$\lambda$ & $0, 0.01, 0.02, \ldots, 0.5$ \\
\bottomrule
\end{tabular*}
\label{table:parameter_values}
\end{table}

\begin{figure}%[!ht]
    \centering
    \includegraphics[width=0.8\textwidth]{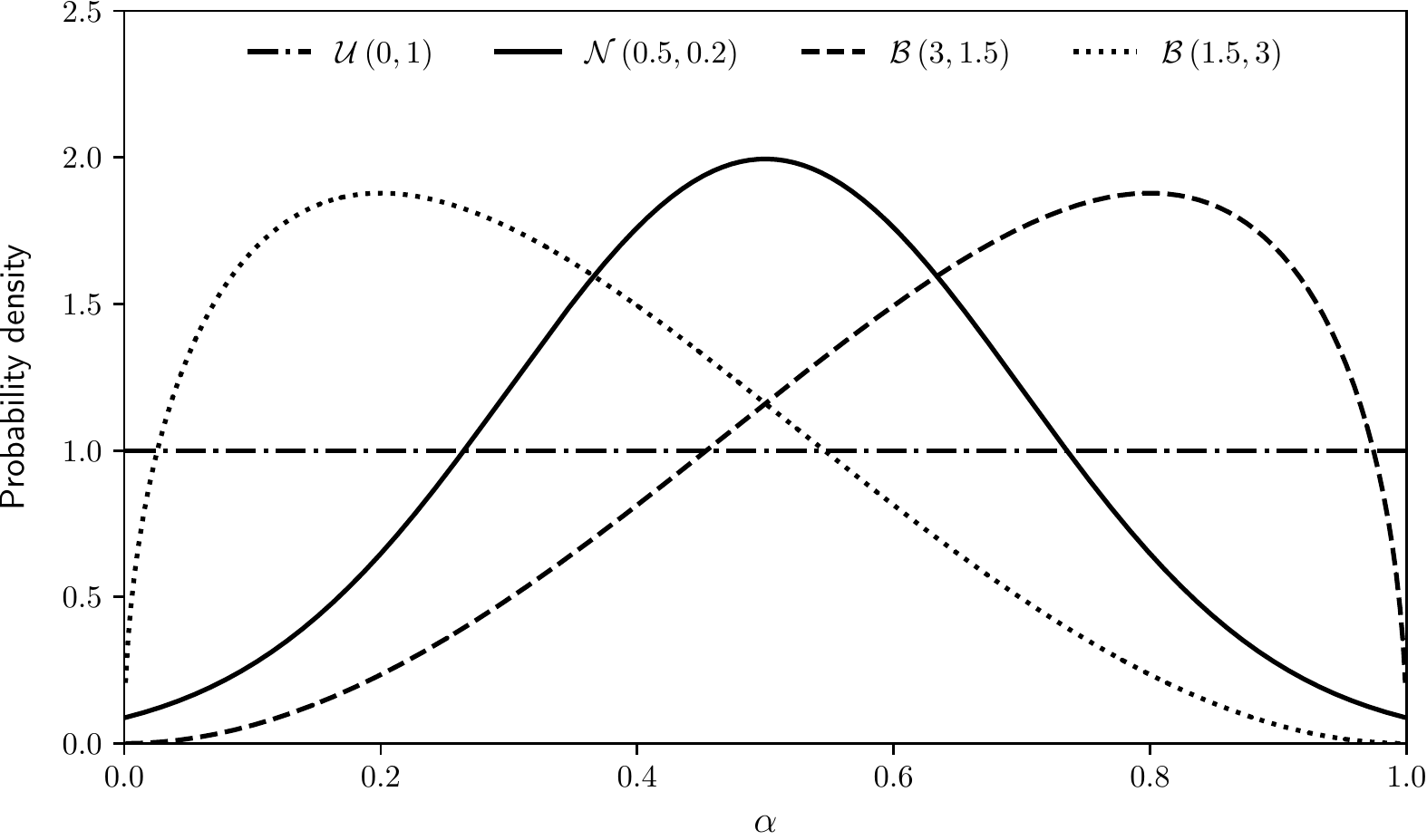}
    \caption{Different probability distributions to generate $\alpha$ values.}
    \label{fig:probability_density}
\end{figure}

For WSM, to generate $\alpha$ values, we have chosen probability distributions in such a way that some ideas could be tested (Figure~\ref{fig:probability_density}). Firstly, the most natural idea is to use a uniform distribution $\mathcal{U}(0,1)$ that generates values between 0 and 1 with the same probability. Using a normal distribution with mean 0.5 and standard deviation 0.2, denoted as $\mathcal{N}(0.5, 0.2)$, we lay emphasis on generating values close to 0.5 in order to focus on weighted-sum objective functions equivalent to the original TTP objective function. Note that the closer to 0.5 the value is, the greater is the interaction between the two components of the problem, and perhaps we should concentrate the algorithm's efforts on these values. On the other hand, maybe we should focus on values close to 0 or 1 when it is the case that one of the components is more easily solved. For example, note that for $\alpha$ values closer to 0, we are looking for TTP solutions with good TSP components (few or no items should be stolen). As we are using the Chained-Lin-Kernighan heuristic, one of the most efficient algorithms for generating near-optimal TSP solutions~\citep{wu2018evolutionary}, our algorithm might not need to exploit these values much to find good TTP solutions concerning good TSP component. Thus, we can use, for example, a beta distribution $\mathcal{B}(3, 1.5)$ that does not generate many values close to 0. In addition, we have also considered in our experiments a beta distribution $\mathcal{B}(1.5, 3)$ with their parameters swapped concerning the previous distribution to address scenarios where the Chained-Lin-Kernighan heuristic combined with our packing algorithm is able to find good TTP solutions with a high collected profit without the need for a high emphasis on $\alpha$ values close to 1. For a reference on probability distributions, we refer to \cite{krishnamoorthy2016handbook}.

To ensure better performance of the proposed algorithm, we have analyzed the influence of its parameters on different types of instances. More precisely, we have divided all 960 instances into 96 groups and then execute Irace on each of them. Each group contains all 10 instances defined with different sizes of knapsacks. These groups are identified as {\tt XXX\_YY\_ZZZ}, in the same way as we have identified the instances, except for the lack of {\tt WW}. With this approach, we would like to know whether there exist similar behaviors among the best parameter configurations from different groups of instances. As we have selected 96 groups with a large difference in characteristics among them, it is reasonable to think that whether such behaviors exist, they may also apply to unknown instances.

As Irace evaluates the quality of the output of a parameter configuration using a single numerical value, we should use for multi-objective problems some unary quality measure \citep{lopez2016irace}, such as the hypervolume indicator or the $\epsilon$-measure \citep{zitzler2003performance}. In our experiments, we have used the hypervolume indicator. In addition, we have used all Irace default settings, except for the parameter \textit{maxExperiments}, which has been set to 1000. This parameter defines the stopping criteria of the tuning process. We refer the readers to \citep{lopez2016iraceguide} for a complete user guide of Irace package.

\begin{figure*}%[!ht]
    \captionsetup[subfigure]{justification=centering, labelformat=empty}
    \centering
    \subfloat[]{\includegraphics[width=0.45\textwidth]{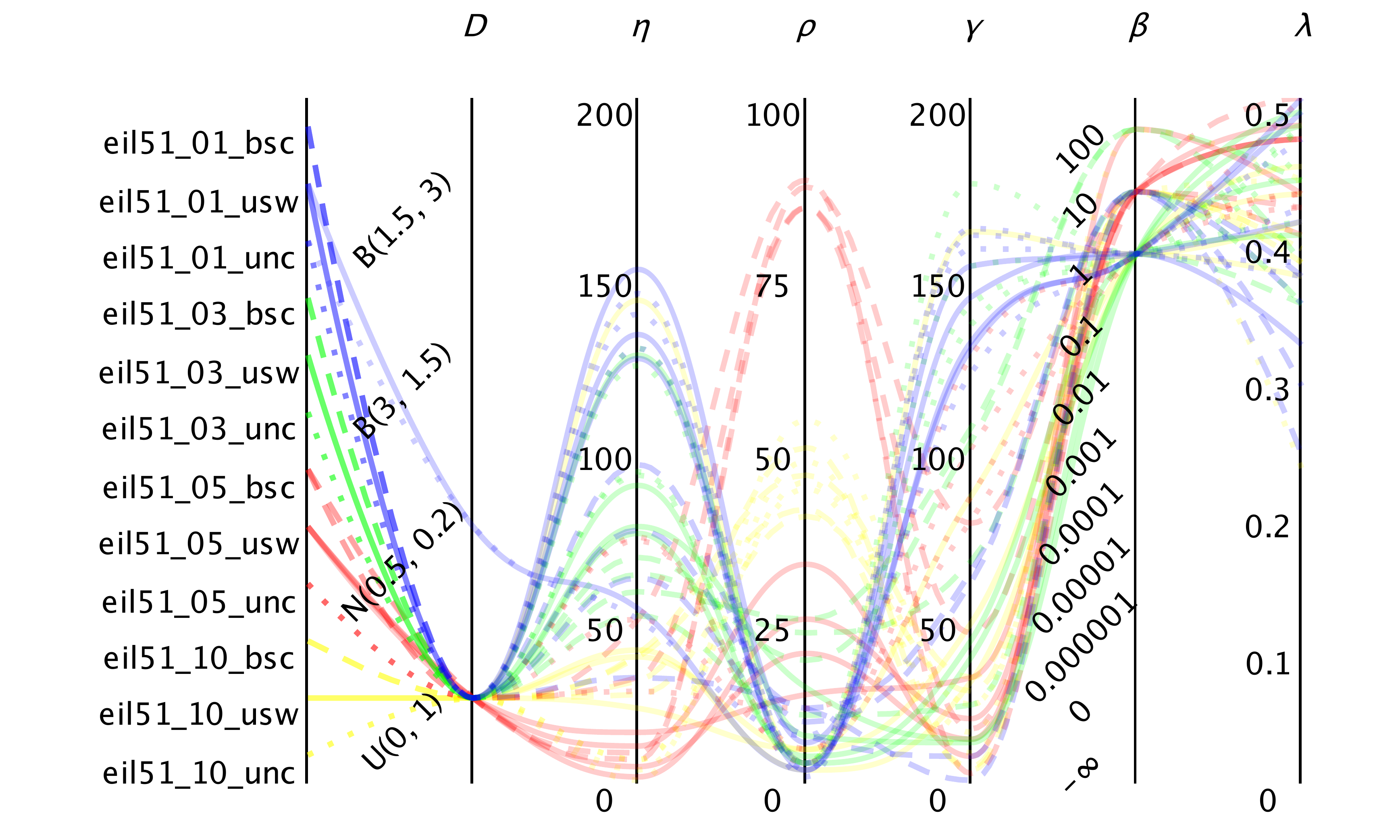}}
    \subfloat[]{\includegraphics[width=0.45\textwidth]{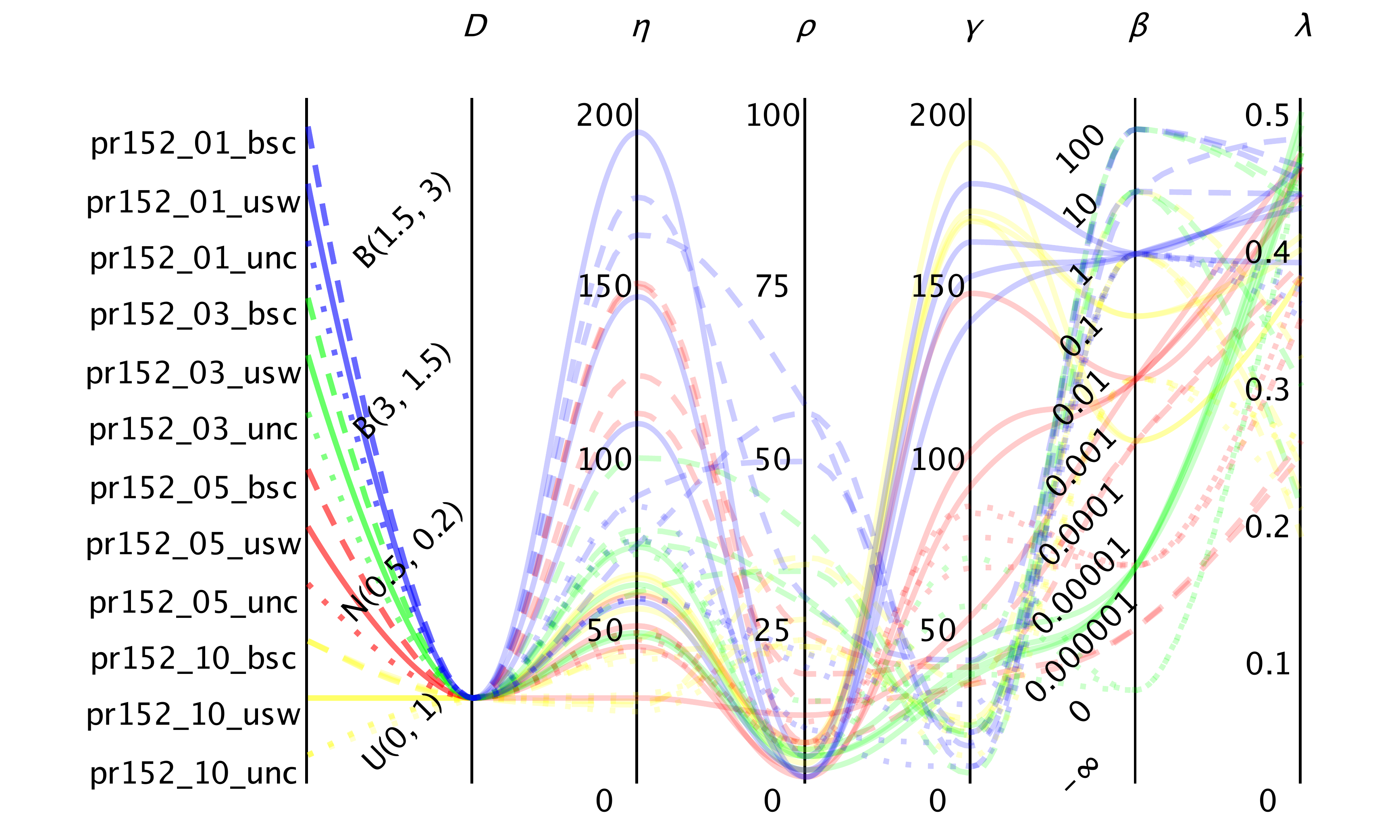}}
    \vspace{-0.8cm}
    \subfloat[]{\includegraphics[width=0.45\textwidth]{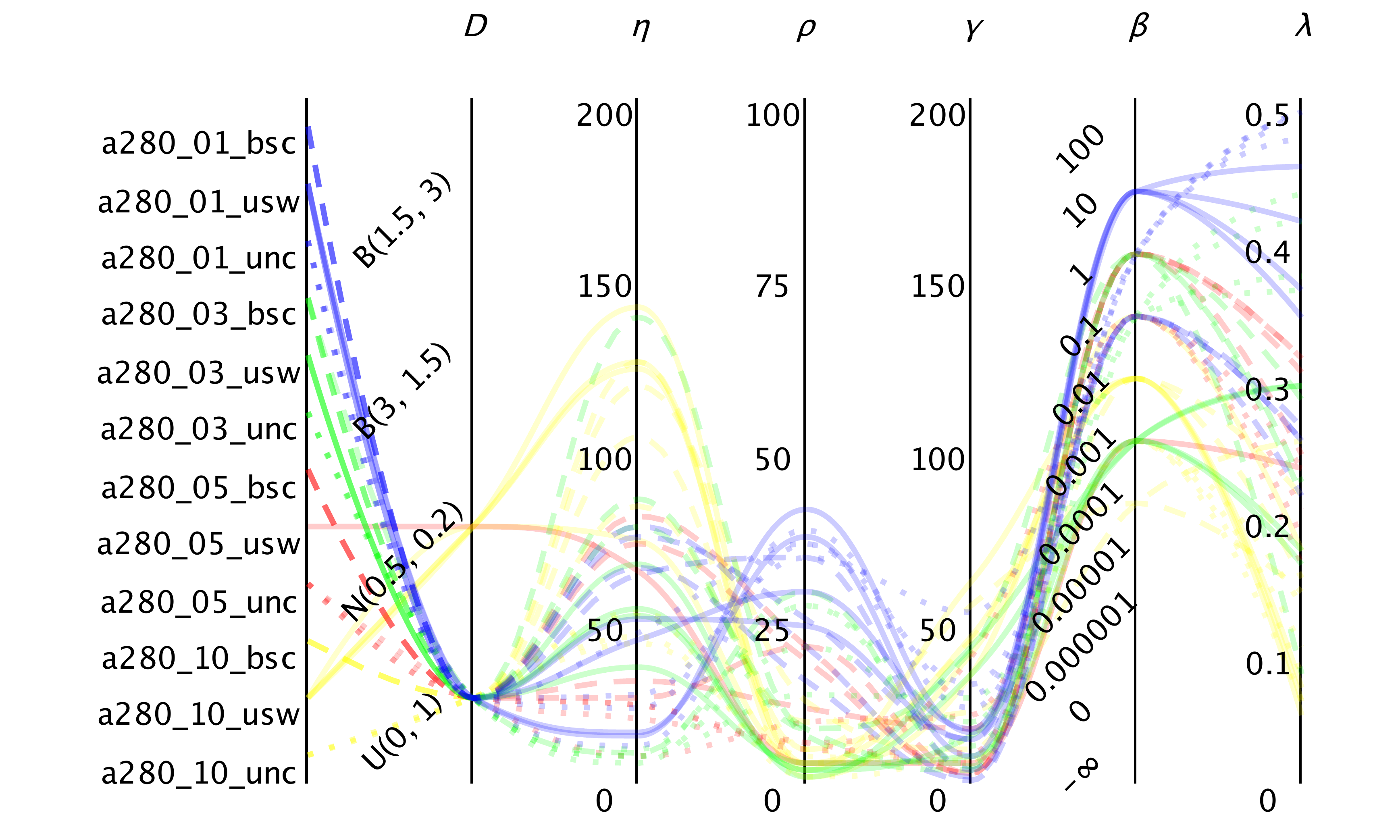}}
    \subfloat[]{\includegraphics[width=0.45\textwidth]{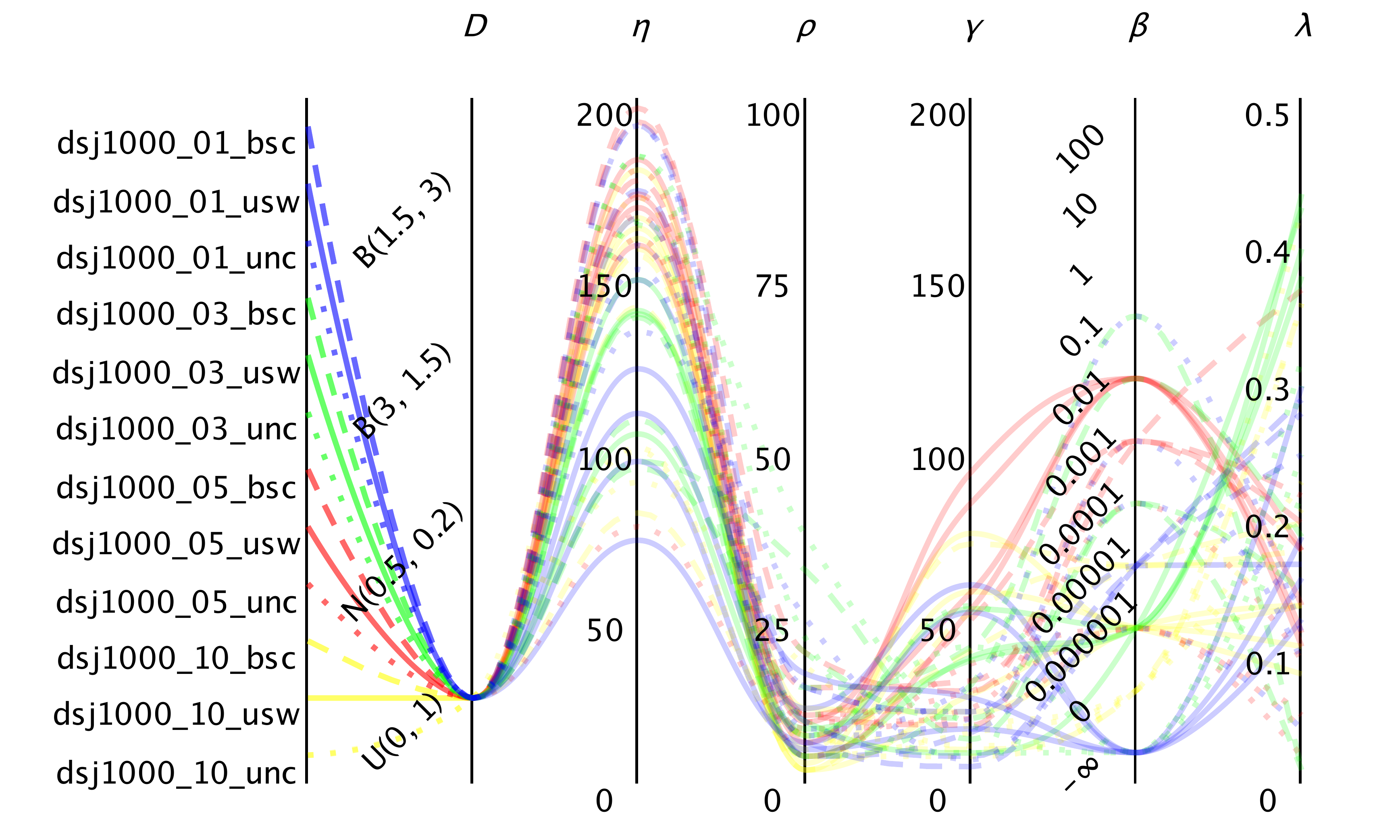}}
    \vspace{-0.8cm}
    \subfloat[]{\includegraphics[width=0.45\textwidth]{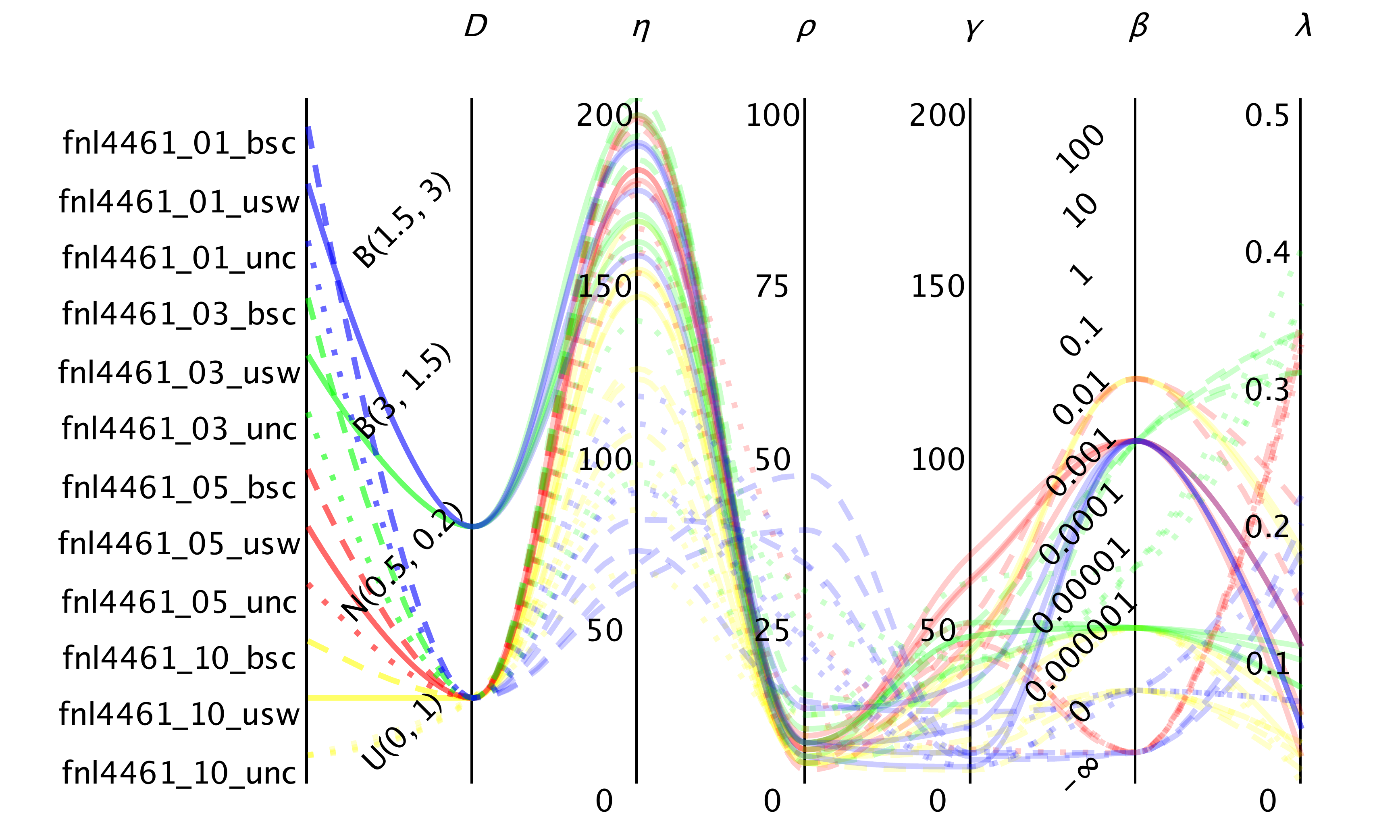}}
    \subfloat[]{\includegraphics[width=0.45\textwidth]{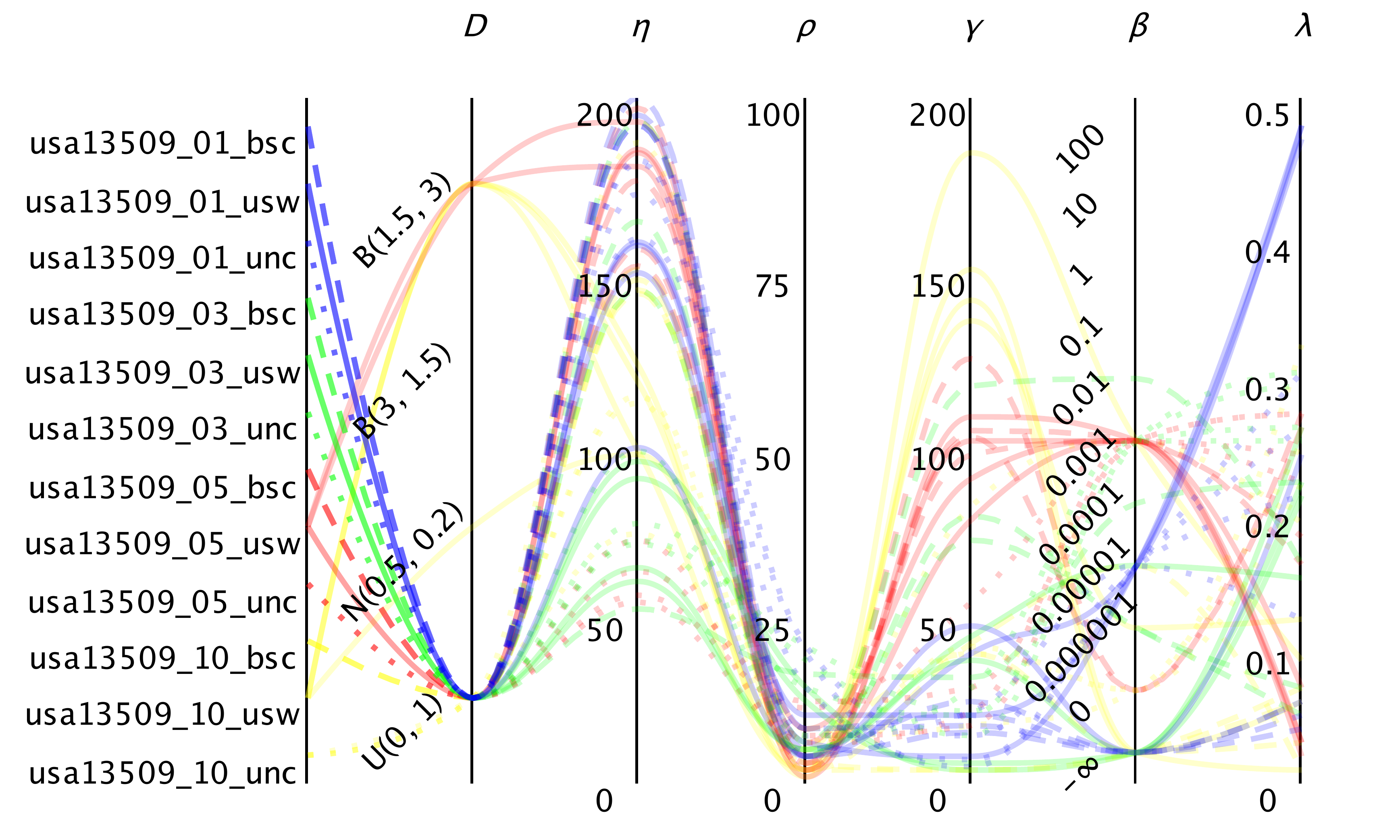}}
    \vspace{-0.8cm}
    \subfloat[]{\includegraphics[width=0.45\textwidth]{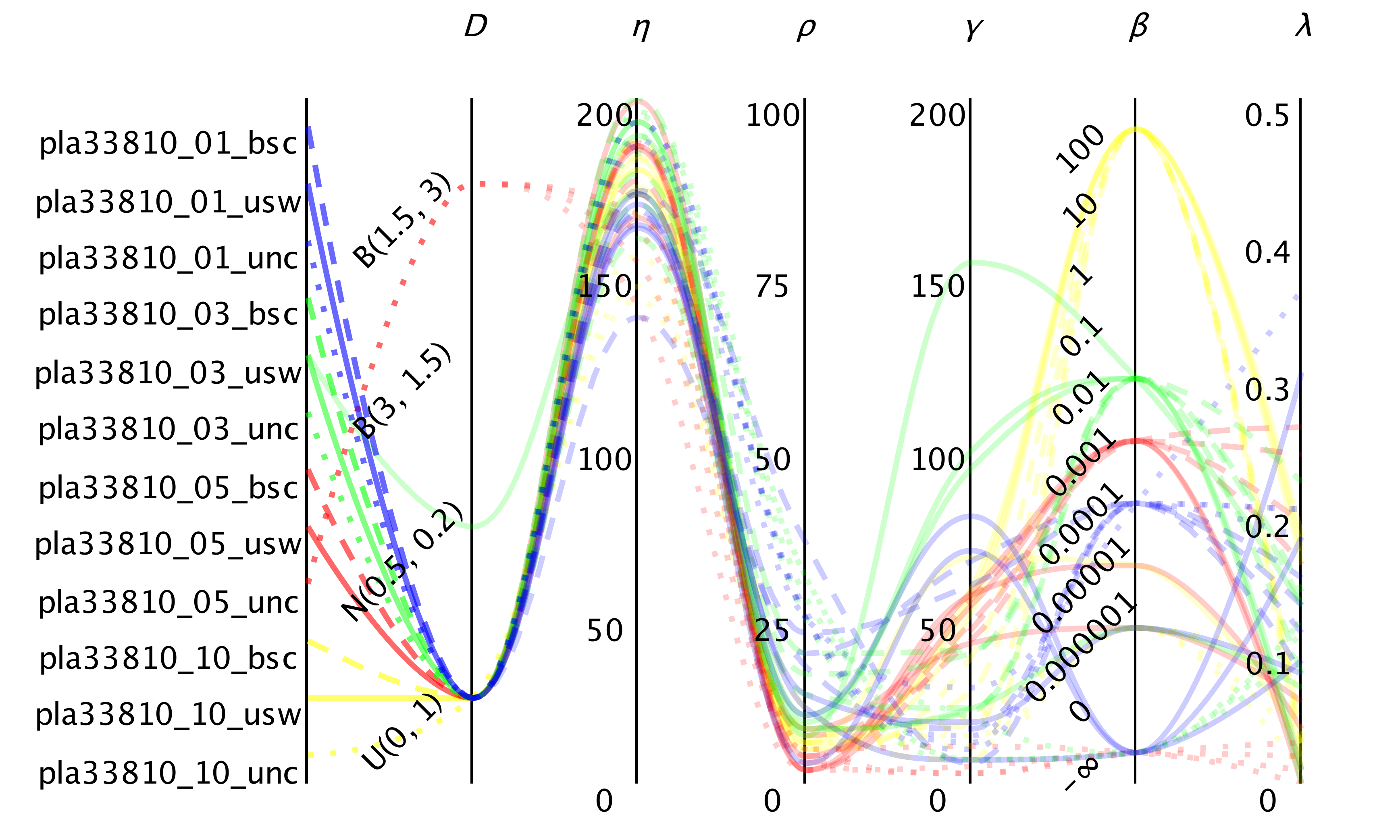}}
    \subfloat[]{\includegraphics[width=0.45\textwidth]{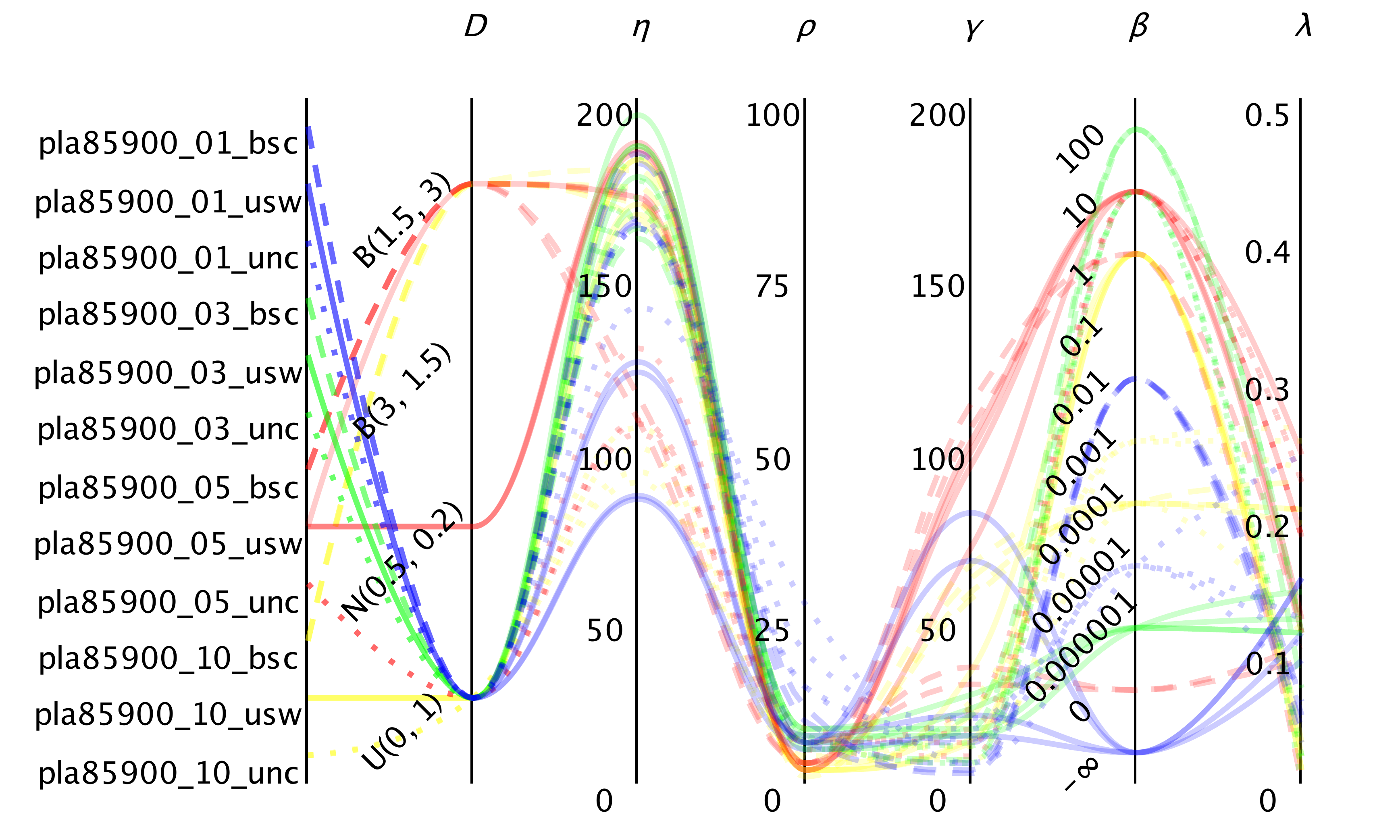}}
    \caption{Irace results for the 96 groups of instances. Blue, green, red, and yellow lines represent, respectively,  groups of instances with 1, 3, 5, and 10 items-per-city. Dashed, solid, and dotted lines are used, respectively, to emphasize the groups of instances with items where their weights and values are bounded and strongly correlated ({\it bsc}), uncorrelated with similar weights ({\it usw}), and uncorrelated ({\it unc}).}
    \label{fig:irace_results}
\end{figure*}

From the tuning experiments, we have obtained the results shown in Figure \ref{fig:irace_results}. Each parallel coordinate plot lists for each of the 96 groups (listed in the left-most column) the configurations returned by Irace (plotted in the other columns). As Irace can return more than one configuration that are statistically indistinguishable given the threshold of the statistical test, multiple configurations are sometimes shown. Each vertical axis indicates a parameter and its range of values, and each configuration of parameters is described by a line that cuts each parallel axis in its corresponding value. Through the concentration of the lines, we can see which parameter values have been most selected among all tuning experiments. We have used different colors and styles for lines in order to emphasize the results obtained for each group individually. All logs generated by the Irace executions, as well as their settings can be found at the GitHub link along with our code.

We can make several observations from the tuning results. First, we notice that for almost all groups of instances the uniform distribution $\mathcal{U}(0,1)$ has been chosen. For some groups, especially those that contain larger instances, other distributions have been returned by Irace. Regarding the parameter $\eta$, we can observe a strong trend in increasing its value as the number of cities increases. This is not too surprising, as the Chained-Lin-Kernighan heuristic, in general, requires more computational time to address larger TSP instances. Thus, computing a higher number of packing plans from each tour may generate better BITTP solutions than resolve the TSP component many times. We can also observe from the values obtained for the parameter $\rho$ that only a few attempts of our packing strategy are needed to reach good results, which is especially true for larger instances. The low values obtained for the parameter $\gamma$ for most groups of instances indicate that the frequency of re-computation of the objective function in the packing algorithm may begin with low values without interfering in the quality of the packing plan computed. Although the values of the parameter $\beta$ do not follow a clear trend, they are strongly related to the number of cities and mainly to the layout that the cities are arranged. For example, when many cities are uniformly arranged, the trend is towards low $\beta$ values as, for this scenario, higher $\beta$ values would probably not be efficient, since the algorithm would spend most of the time processing too many tours obtained from 2-opt moves. Finally, we can see that, in general, higher $\lambda$ values are concentrated in smaller-size instances, which is not surprising since the bit-flip operator would perform too many moves on instances with many items and higher $\lambda$ values.

With a closer look, we can make additional observations by combining different parameters and characteristics of the instances. For example, $\eta$ values are low for medium and large knapsack capacities (red and yellow) of eil51, while the opposite is true for the dsj1000 instances, and other large instances. Across almost all instances, the $\eta$ values are the lowest or among the lowest for instances with uncorrelated (dotted) knapsacks. 
For $\rho$, it is difficult to extract patterns, however, we can observe that the tuned configurations for instances with strongly correlated knapsacks (dashed) have the highest $\rho$ values for the groups eil51 (red), pr152 (blue), and fnl4661 (blue). 
For $\gamma$, small knapsacks (blue) with uncorrelated but similar weights (solid) result in high or the highest values for eil51, pr152, and pla33810, but for example not for a280.
For $\beta$, we cannot observe clear trends for the knapsack type, however, sometimes the knapsack capacity stands out. For example, for a280, the smallest knapsacks (blue) resulted in the highest values, while blue has the lowest values for dsj1000, and the largest knapsacks resulted in the highest  values (yellow) for the tuning experiments for the pla33810 group. 
We can observe similar `inversions' also for $\gamma$. There, for example the smallest knapsacks with uncorrelated and similar weights (blue, solid) result in the smallest values on some instance groups, but for the largest values on others.

In summary, we can observe many consistent as well as inconsistent patterns for the different groups of instances, and depending on the knapsack type and the knapsack capacity. In combination with instance features (e.g. the ones from~\cite{wagner2018case}), this might make for an interesting challenge for per-instance-algorithm-configuration~\citep{Hutter2006piac}, however, this is beyond the scope of the present study. 

\begin{figure*}%[!ht]
    \captionsetup[subfigure]{justification=centering, labelformat=empty}
    \centering
    \subfloat[]{\includegraphics[width=0.95\textwidth]{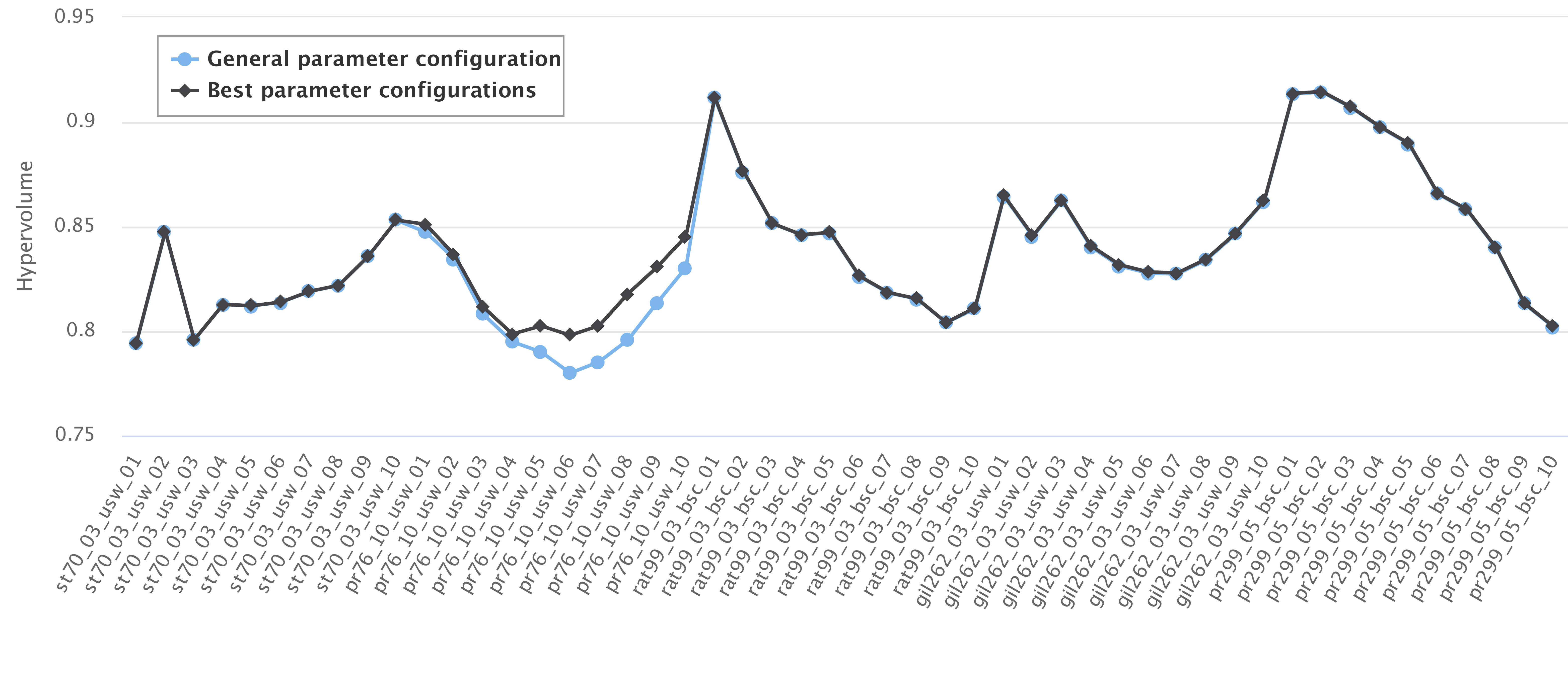}}
    \vspace{-0.8cm}
    \subfloat[]{\includegraphics[width=0.95\textwidth]{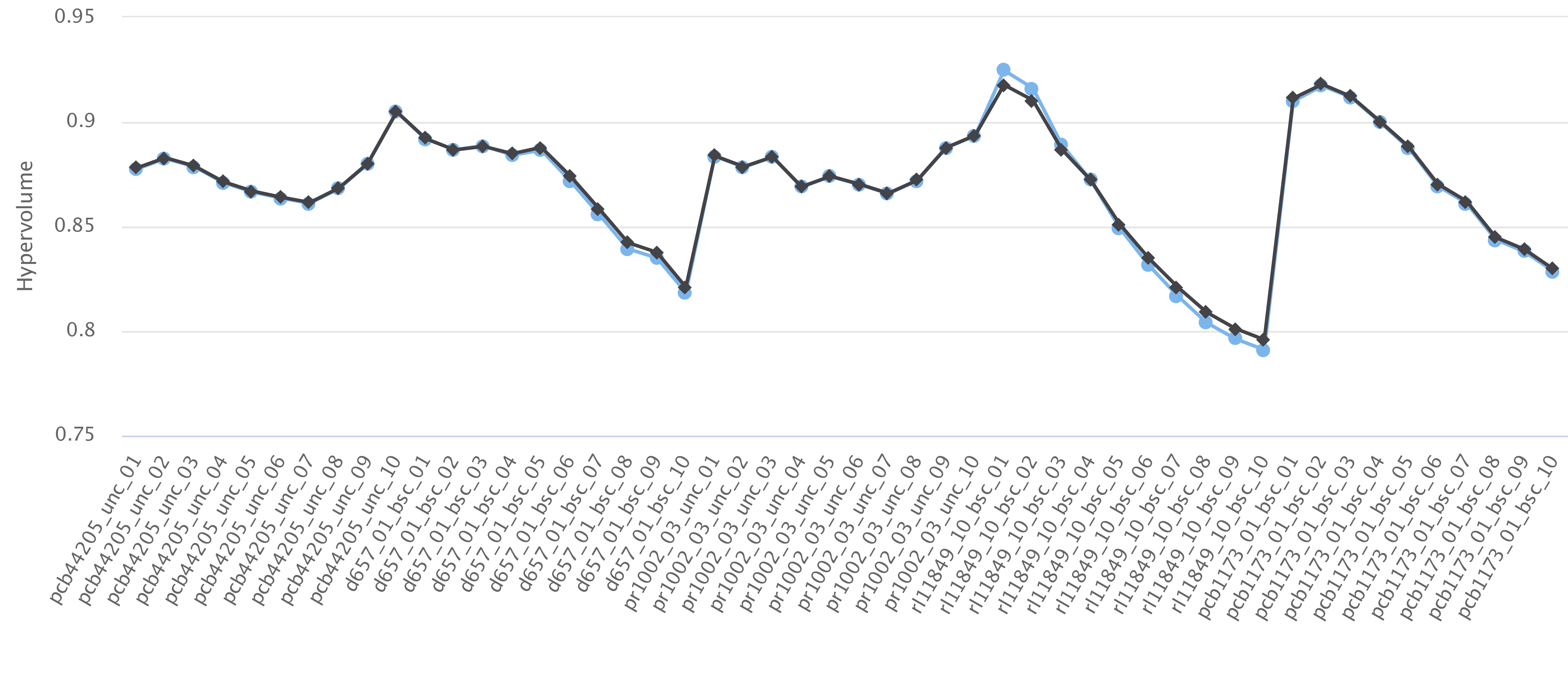}}
    \caption{Average hypervolume on 100 unseen instances. Shown are the results for the best (irace-tuned) parameter configuration (per instance) and of the general (``averaged'') parameter configuration from earlier experiments.}
    \label{fig:parameter_validation}
\end{figure*}

In a final experiment, we investigate the extent to which the results so far can carry over to unseen instances. To achieve this, we use the average of the parameter values (and the mode for the categorical parameters) obtained from the tuning experiments to furnish a single configuration of parameters for unknown instances, and then investigate that configuration's performance. The configuration is: $\mathcal{D}$~=~$\mathcal{U}(0,1)$, $\eta$~=~117, $\rho$~=~12, $\gamma$~=~41, $\beta$~=~0.001, and $\lambda$~=~0.22. In order to confirm whether these values are a reasonable configuration of parameters for our algorithm, we have randomly chosen 10 unknown (from the perspective of the tuning experiments) groups of instances from the instances defined by~\cite{polyakovskiy2014comprehensive}, and compared the results obtained with the aforementioned configuration against the best configuration for each group. To find out which is the best configuration of parameters for each of these 10 groups, we have again used Irace in the same way as described before. As each group has 10 instances, we have 100 instances in total. For each of them and for each of the two different configurations of parameters, we have run our algorithm 30 times and used average values of hypervolumes achieved to plot Figure~\ref{fig:parameter_validation}. There, two sets stand out: (i) on the pr76 instances the  tuned configuration performs better; and (ii) on the rl11849 the general parameter configuration is sometimes worse and sometimes better. 
On all other instances, the different configurations (general \textit{vs.} best configuration) perform essentially the same.

\subsection{WSM \textit{vs.} NDSBRKGA}

In the first analysis that assesses the quality of our WSM, we compare the solutions obtained by it with the solutions obtained by NDSBRKGA proposed by \cite{chagas2020nondominated}. In order to make a fair comparison, we have tuned the parameters of the NDSBRKGA following the same procedure used in the tuning of the parameters of WSM. The parameter values considered for these experiments have been chosen based on the insights reported in~\citep{chagas2020nondominated}. These parameter values, as well as the results obtained in the experiment, are available at the GitHub link along with our other files.\footnote{As neither the HPI algorithm nor the HPI implementation are available, we could not include HPI in this tuning-based comparison. HPI has been the result of a classroom setting and their actual results have been (to some extent) aggregated across multiple teams (and hence implementations), which has been legal w.r.t. the competition rules (as said competition required only the solution files, not any implementations).}

Due to the randomized nature of both algorithms, we have performed 30 independent repetitions on each instance. Each run has been executed for 10 minutes with the best parameter values found in the tuning experiments.

As in \citep{chagas2020nondominated}, we have used the hypervolume indicator (HV) \citep{zitzler1998multiobjective} as a performance indicator to compare and analyze the results obtained. This indicator is one of the most used indicators for measuring the quality of a set of non-dominated solutions by calculating the volume of the dominated portion of the objective space bounded from a reference point. To make the hypervolume suitable for the comparison of objectives with greatly varying ranges, a normalization of objective values is commonly done beforehand. Therefore, before computing the hypervolume, we have first normalized the objective values between 0 and 1 according to their minimum and maximum value found during our experiments. Although maximizing the hypervolume might not be equivalent to finding the optimal approximation to the Pareto-optimal front \citep{BRINGMANN2013265,WAGNER2015465}, we have assumed that the higher the hypervolume indicator, the better the solution sets are, as is commonly considered in the literature.

We compare the performance of the solutions obtained by measuring for each instance the percentage variation of the average hypervolume obtained considering the independent runs of each algorithm. More precisely, for each instance, we have estimated the reference point as the maximum travel time and the minimum profit obtained from the non-dominated solutions, which have been extracted from all solutions returned by the algorithms. Then, we have computed the hypervolume covered by the non-dominated solutions found by each run of each algorithm according to the estimated reference point. Thereafter, we can compute the percentage variation as
\begin{equation}
    \big(\text{HV}^{\text{WSM}}_{\text{avg}} - \text{HV}^{\text{NDSBRKGA}}_{\text{avg}}\big)\;/\; \text{max}\big(\text{HV}^{\text{WSM}}_{\text{avg}}, \text{HV}^{\text{NDSBRKGA}}_{\text{avg}}\big) \cdot 100\% \nonumber
\end{equation}
\noindent, where $\text{HV}^{\text{WSM}}_{\text{avg}}$ and $\text{HV}^{\text{NDSBRKGA}}_{\text{avg}}$ are, respectively, the average hypervolumes obtained by WSM and NDSBRKGA in their independent executions.

\begin{figure}%[!ht]
    \centering
    \includegraphics[width=0.82\textwidth]{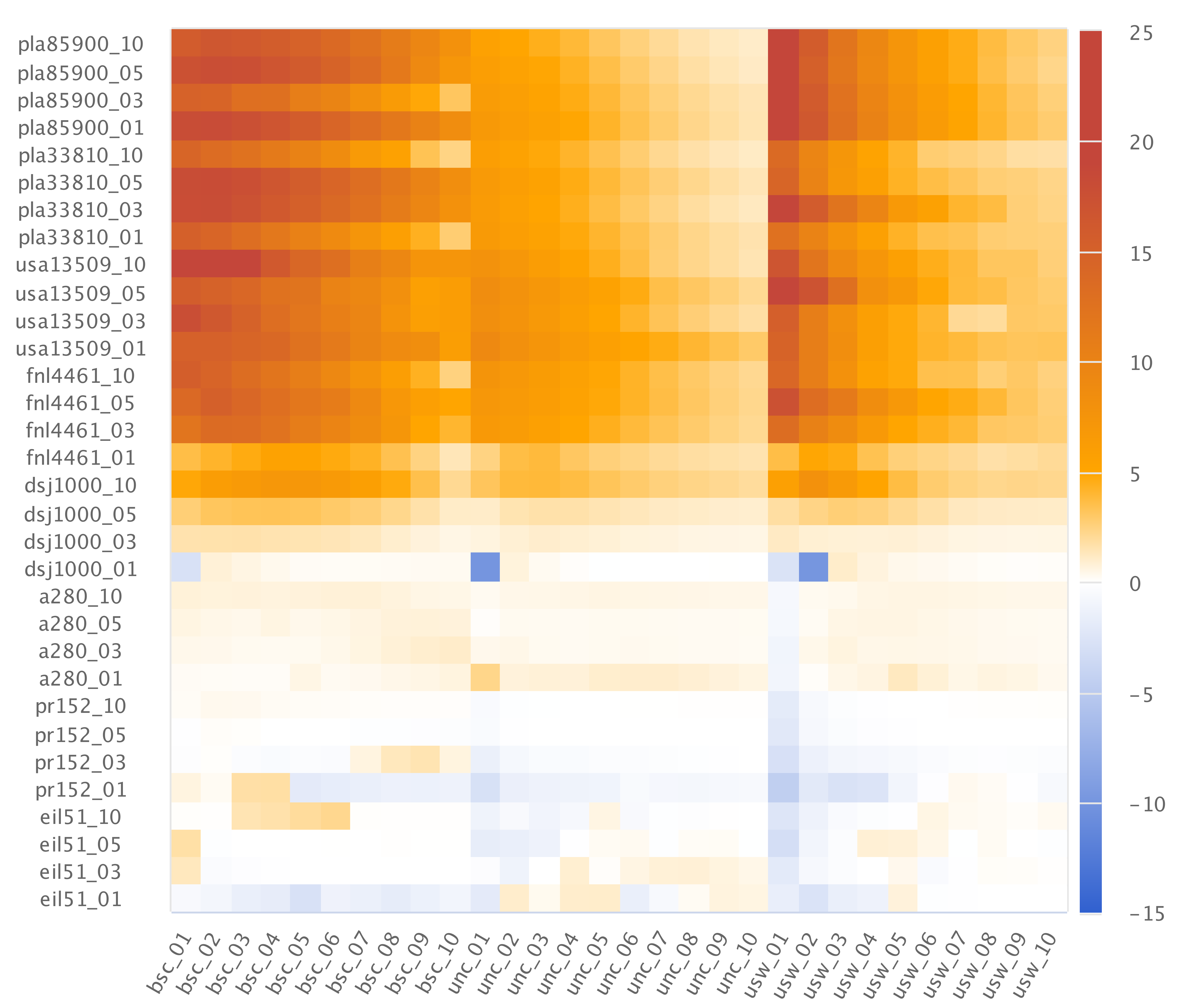}
    \caption{Percentage variation of the average hypervolumes. Shades of orange and red indicate in which instances our WSM has reached a higher hypervolume than NDSBRKGA, while shades of blue indicate the opposite.}
    \label{fig:hv_wsm_vs_ndsbrkga}
\end{figure}

In Figure \ref{fig:hv_wsm_vs_ndsbrkga}, we visualise the percentage variations of the average hypervolumes using a heatmap to emphasize larger variations. Each cell of the heatmap informs the results obtained for a specific instance. Note that the vertical axis depicts the characteristics {\tt XXX} and {\tt YY} of instances, while the horizontal axis depicts the characteristics {\tt ZZZ} and {\tt WW}. Note also that positive variation values (highlighted in shades of orange and red) indicate that our WSM has reached a higher hypervolume, while negative variation values (highlighted in shades of blue) indicate the opposite behavior. Besides, the higher the absolute value (more intense color), the higher the difference between the hypervolumes.

From Figure~\ref{fig:hv_wsm_vs_ndsbrkga}, we can observe that our WSM is clearly more effective than NDSBRKGA for larger instances. This is especially true for instances with the smallest knapsack capacities. Note that, in general, the smaller the size is of the knapsack, the higher the performance is of WSM concerning the NDSBRKGA.

Note that, although our solutions still cover a higher hypervolume for larger instances with uncorrelated ({\it unc}) items, it must be stressed that our WSM has obtained the worse performance regarding the NDSBRKGA for these instances. This behavior could be explained by the fact that our packing heuristic might present difficulties in dealing with those items because when there is no correlation between their profits and weights and the weights present a large variety, our packing algorithm may not be able to create a good order of the items for our packing strategy. This is interesting, as \textit{unc} knapsacks are not necessarily seen as difficult \citep{martello1999dynamic}; but in our algorithm, they might end up being due to our strategy for solving the KP component. Another fact that could explain the worse performance of WSM on instances with uncorrelated items would be that NDSBRKGA has a good performance for these instances, making the performance of our algorithm less prominent.

\begin{figure*}%[!ht]
    \captionsetup[subfigure]{justification=centering}%, labelformat=empty}
    \centering
    \subfloat[NDSBRKGA]{\includegraphics[width=0.70\textwidth]{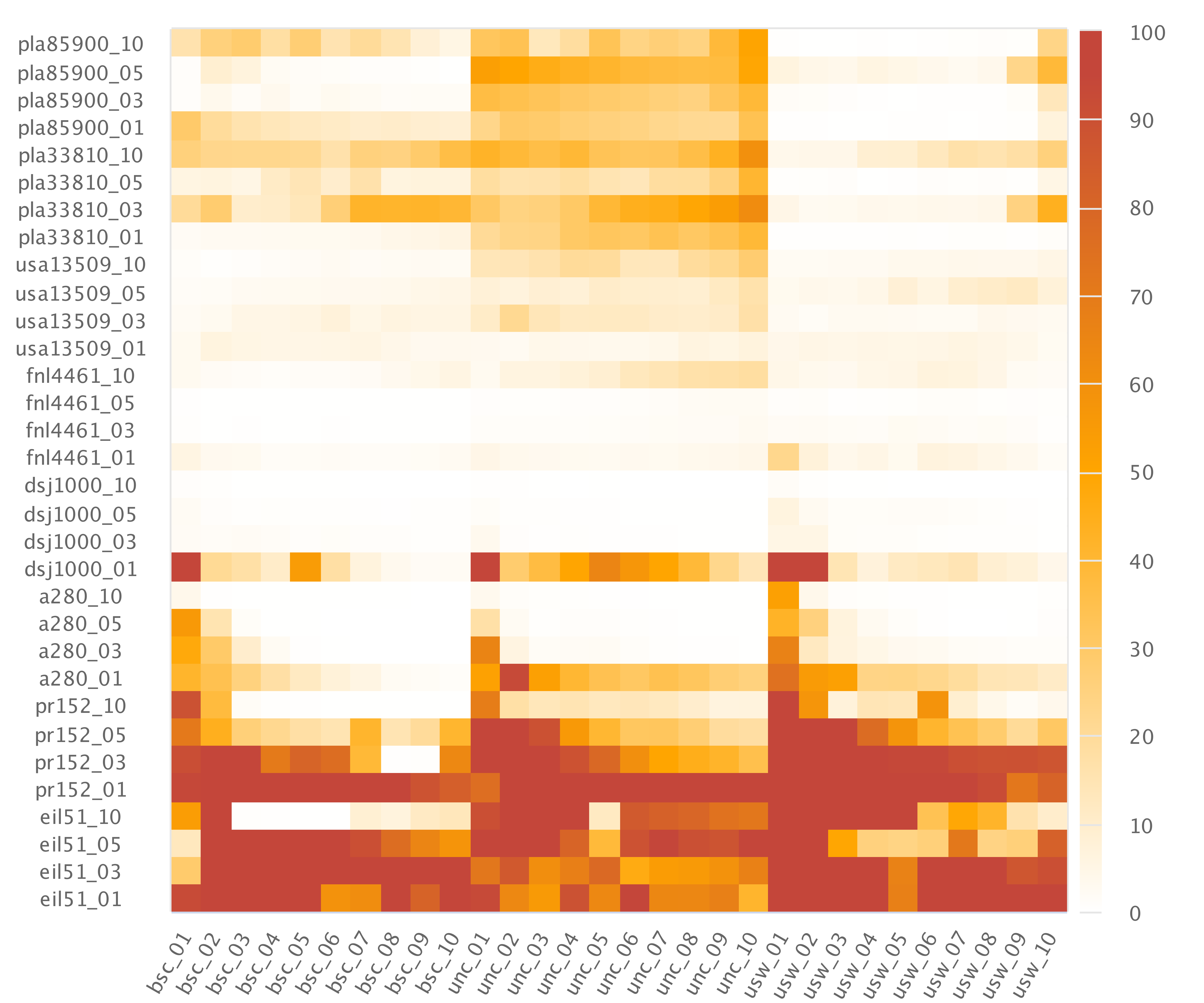}}
    
    \subfloat[WSM]{\includegraphics[width=0.70\textwidth]{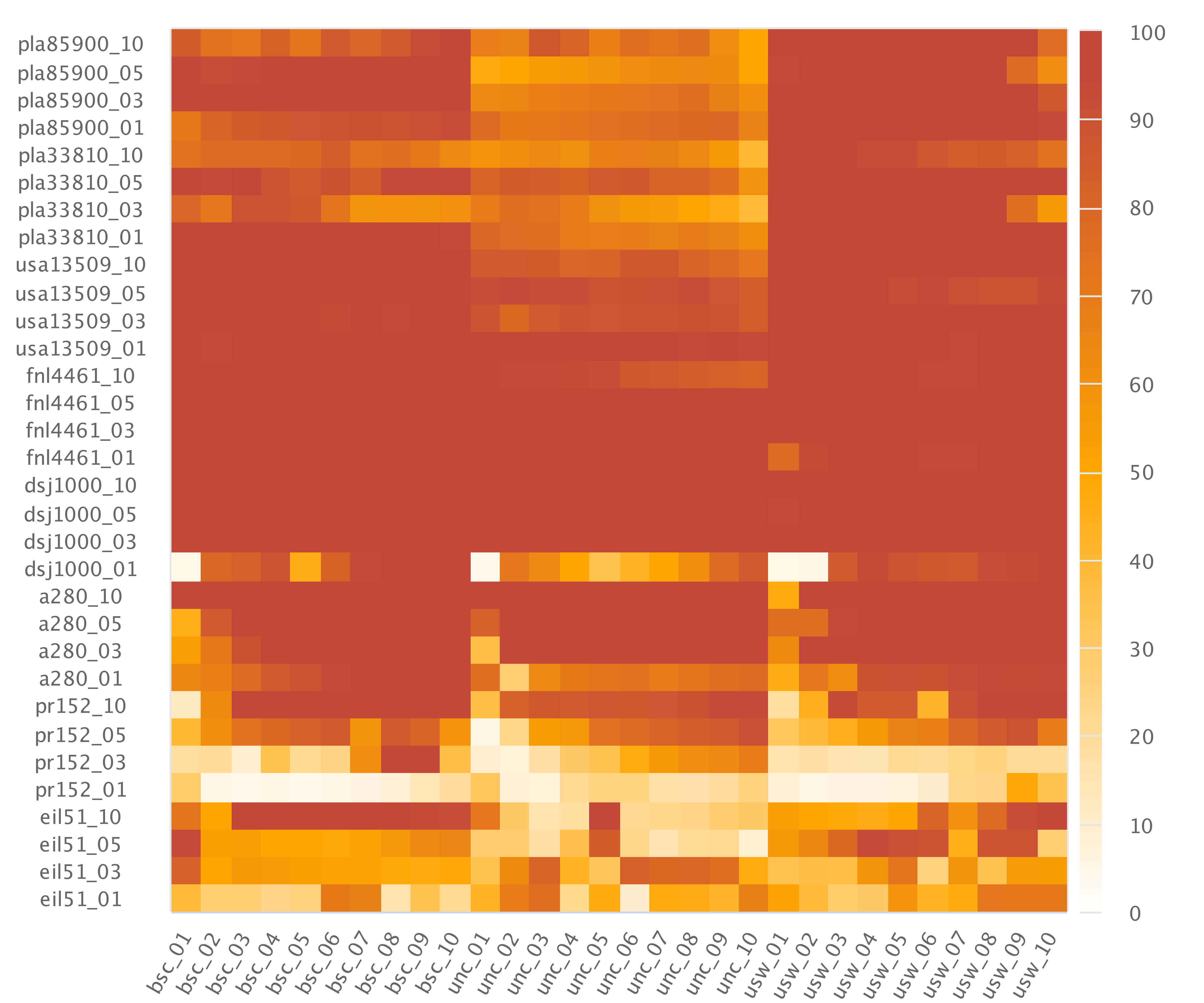}}
    \caption{Percentage of non-dominated solutions found by each algorithm.}
    \label{fig:percentage_number_nds}
\end{figure*}

Regarding the smaller-size instances, both algorithms have achieved similar performance (almost blank cells). However, with a closer look at Figure~\ref{fig:hv_wsm_vs_ndsbrkga}, we can see a slightly better performance of NDSBRKGA. To better analyze these results, we have used another performance measure. For each instance we have merged all the solutions found in order to extract from them a single non-dominated set of solutions. Then, we have computed how many non-dominated solutions have been obtained by each algorithm. Our purpose of this analysis is to evaluate both algorithms regarding their ability to find non-dominated solutions with different objective values. Therefore,  duplicate solutions regarding their objective values have been removed, i.e., we have regarded a single non-dominated solution with the same values in both objectives. In Figure~\ref{fig:percentage_number_nds}, we present these numbers in percentages according to the total number of non-dominated solutions following the heatmap scheme used previously. 

The results shown in Figure~\ref{fig:percentage_number_nds} corroborate those shown in Figure~\ref{fig:hv_wsm_vs_ndsbrkga}. As was expected, our algorithm has found more non-dominated solutions especially for those instances where it obtained a higher hypervolume. However, even for the instances in which the NDSBRKGA found better solutions, the difference between the hypervolumes of both algorithms remains low.

To statistically compare the performance of the algorithms, we have used the Wilcoxon signed-rank test on the hypervolumes achieved in the 30 independent runs. With a significance level of 5\%, there is no statistical difference between both algorithms in 27 instances (2.8\%), our algorithm is significantly better in 789 instances (82.2\%) and worse in 144 (15\%) ones when compared to NDSBRKGA. 

\subsection{WSM \textit{vs.} competition results}

Next, we compare WSM to the results of the BITTP competitions held at EMO2019\footnote{\url{https://www.egr.msu.edu/coinlab/blankjul/emo19-thief/}} and GECCO2019\footnote{\url{https://www.egr.msu.edu/coinlab/blankjul/gecco19-thief/}}. Both competitions have used the same rules and criteria. There were no regulations regarding the running time and the number of processors used. The final ranking used for the competitions was solely based on the solution set submitted by each participant for nine medium/large TTP instances chosen from the TTP benchmark \citep{polyakovskiy2014comprehensive}. More precisely, the final ranking was defined according to the hypervolume covered by the solutions. To calculate the hypervolumes, the reference points have been defined as the maximum time and the minimum profit obtained from the non-dominated solutions, which have been built from all submitted solutions. In order to make a fair ranking, the maximum number of solutions allowed for each instance has been limited. In Table~\ref{table:competition_instances}, we list the instances used as well as the maximum number of solutions allowed.

\begin{table}[!ht]
%\tiny
%\scriptsize
%\footnotesize
%\small
%\normalsize
\centering
\caption{Maximum number of solutions allowed by each TTP instance used in the BITTP competitions.}\small
\setlength{\tabcolsep}{0pt}
\begin{tabular*}{\hsize}{@{}@{\extracolsep{\fill}}cc@{}}
\toprule
\multicolumn{1}{c}{Instance} & \begin{tabular}[x]{@{}c@{}}Maximum number of\\ solutions allowed\end{tabular} \\
\midrule
a280\_01\_bsc\_01 & 100 \\ 
a280\_05\_usw\_05 & 100 \\ 
a280\_10\_unc\_10 & 100 \\[1mm]
fnl4461\_01\_bsc\_01 & 50 \\ 
fnl4461\_05\_usw\_05 & 50 \\ 
fnl4461\_10\_unc\_10 & 50 \\[1mm]
pla33810\_01\_bsc\_01 & 20 \\ 
pla33810\_05\_usw\_05 & 20 \\ 
pla33810\_10\_unc\_10 & 20 \\ 
\bottomrule
\end{tabular*}
\label{table:competition_instances}
\end{table}

As our algorithm can return a higher number of solutions than those reported in Table~\ref{table:competition_instances}, we have used the dynamic programming algorithm developed by \cite{auger2009investigating} in order to find a subset of limited size of the returned solutions such that their hypervolume indicator is maximal. As stated by \cite{auger2009investigating}, this dynamic programming can be solved in time $\mathcal{O}(\vert A \vert ^{3})$, where $A$ would be the set of solutions returned by our algorithm. Note that the application of this strategy has also been used in \citep{chagas2020nondominated} for NDSBRKGA, and it is only part of a post-processing needed to fit both algorithms to the competition criteria. 

In both competitions, preliminary versions of NDSBRKGA have been submitted as \textit{jomar}, a reference to the two authors (\textbf{Jo}natas and \textbf{Mar}cone) who first worked on that algorithm. These preliminary versions are presented in \citep{chagas2020nondominated} as well as their results achieved in both competitions. In short, \textit{jomar} has won the first and second places, at EMO2019 and GECCO2019 competitions, respectively. After the competitions, some improvements have been incorporated in the preliminary versions of NDSBRKGA, resulting in its final version as is described  in \citep{chagas2020nondominated}. In the following, we compare our WSM with that final version as it presents slightly better results concerning its previous ones. 

In Table~\ref{table:competition_results}, we present for each instance the best results submitted for the competitions and also the results obtained by our WSM. The results of all submissions can be found at web pages previously reported. As the final results of NDSBRKGA have been obtained with 5 hours of processing, we have executed our algorithm for 5 hours as well to make a fair comparison. We would like to mention that we have no information on how the other participants have obtained their results. As we stated before, there were no regulations regarding the running time and the number of processors used. In both competitions, their rankings have been solely based on the solution set submitted by each participant. Furthermore, to the best of our knowledge, there is no description available of the solution approaches submitted.

\begin{table}%[!ht]
\renewcommand{\arraystretch}{0.52}
%\tiny
%\scriptsize
\footnotesize
%\small
%\normalsize
\centering
\caption{Best BITTP competitions results \textit{vs.} WSM.}\small
\setlength{\tabcolsep}{0pt}
\begin{tabular*}{\hsize}{@{}@{\extracolsep{\fill}}ccr@{}}
\toprule
\multicolumn{1}{c}{Instance} & \multicolumn{1}{c}{Participant/Algorithm} & \multicolumn{1}{c}{HV} \\ 
\midrule
\multirow{4}{*}{a280\_01\_bsc\_01} & HPI & 0.898433 \\ 
 & NDSBRKGA & 0.895708 \\ 
 & \textBF{WSM} & \textBF{0.887205} \\ 
 & shisunzhang & 0.886576 \\ 
% & NTGA & 0.883706 \\
\midrule
\multirow{4}{*}{a280\_05\_usw\_05} & NDSBRKGA & 0.826879 \\ 
 & HPI & 0.825913 \\ 
 & shisunzhang & 0.820893 \\ 
 & \textBF{WSM} & \textBF{0.820216} \\ 
% & NTGA & 0.811490 \\
\midrule
\multirow{4}{*}{a280\_10\_unc\_10} & NDSBRKGA & 0.887945 \\ 
  & \textBF{WSM} & \textBF{0.887680} \\ 
  & HPI & 0.887571 \\ 
  & ALLAOUI & 0.885144 \\ 
%  & NTGA & 0.882562 \\
\midrule
\multirow{4}{*}{fnl4461\_01\_bsc\_01} & \textBF{WSM} & \textBF{0.934685} \\
 & NDSBRKGA & 0.933942 \\ 
 & HPI & 0.933901 \\ 
 & NTGA & 0.914043 \\ 
% & ALLAOUI & 0.889219 \\
\midrule
\multirow{4}{*}{fnl4461\_05\_usw\_05} & \textBF{WSM} & \textBF{0.820481} \\
 & HPI & 0.818938 \\ 
 & NDSBRKGA & 0.814492 \\ 
 & NTGA & 0.803470 \\ 
% & SSteam & 0.781462 \\
\midrule
\multirow{4}{*}{fnl4461\_10\_unc\_10} & \textBF{WSM} & \textBF{0.882932} \\
 & HPI & 0.882894 \\ 
 & NDSBRKGA & 0.874688 \\ 
 & SSteam & 0.856863 \\ 
% & shisunzhang & 0.850339 \\
\midrule
\multirow{4}{*}{pla33810\_01\_bsc\_01} & \textBF{WSM} & \textBF{0.930580} \\
 & HPI & 0.927214 \\ 
 & NTGA & 0.888680 \\ 
 & ALLAOUI & 0.873717 \\ 
% & NDSBRKGA & 0.852836 \\
\midrule
\multirow{4}{*}{pla33810\_05\_usw\_05} & \textBF{WSM} & \textBF{0.819743} \\
 & HPI & 0.818259 \\ 
 & NDSBRKGA & 0.781009 \\ 
 & SSteam & 0.776638 \\ 
% & NTGA & 0.773589 \\
\midrule
\multirow{4}{*}{pla33810\_10\_unc\_10} & \textBF{WSM} & \textBF{0.876805} \\
 & HPI & 0.876129 \\ 
 & NDSBRKGA & 0.857105 \\ 
 & SSteam & 0.853805 \\ 
% & ALLAOUI & 0.836965 \\
\bottomrule
\end{tabular*}
\begin{tablenotes}
\small
\item \footnotesize\textbf{ALLAOUI} is formed by Mohcin Allaoui and Belaid Ahiod; \textbf{HPI} is formed by Tobias Friedrich, Philipp Fischbeck, Lukas Behrendt, Freya Behrens, Rachel Brabender, Markus Brand, Erik Brendel, Tim Cech, Wilhelm Friedemann, Hans Gawendowicz, Merlin de la Haye, Pius Ladenburger, Julius Lischeid, Alexander Löser, Marcus Pappik, Jannik Peters, Fabian Pottbäcker, David Stangl, Daniel Stephan, Michael Vaichenker, Anton Weltzien, and Marcus Wilhelm; \textbf{NTGA} is formed by Maciej Laszczyk and Pawel Myszkowski; \textbf{shisunzhang} is formed by Jialong Shi, Jianyong Sun, and Qingfu Zhang; and \textbf{SSteam} is formed by Roberto Santana, and Siddhartha Shakya.
\end{tablenotes}
\label{table:competition_results}
\end{table}

\begin{figure*}%[!ht]
    \captionsetup[subfigure]{position=top, justification=centering, labelformat=empty, oneside, margin={0.8cm,0cm}}
    \centering
    \subfloat[\footnotesize\texttt{a280\_01\_bsc\_01}]{\includegraphics[width=0.33\textwidth]{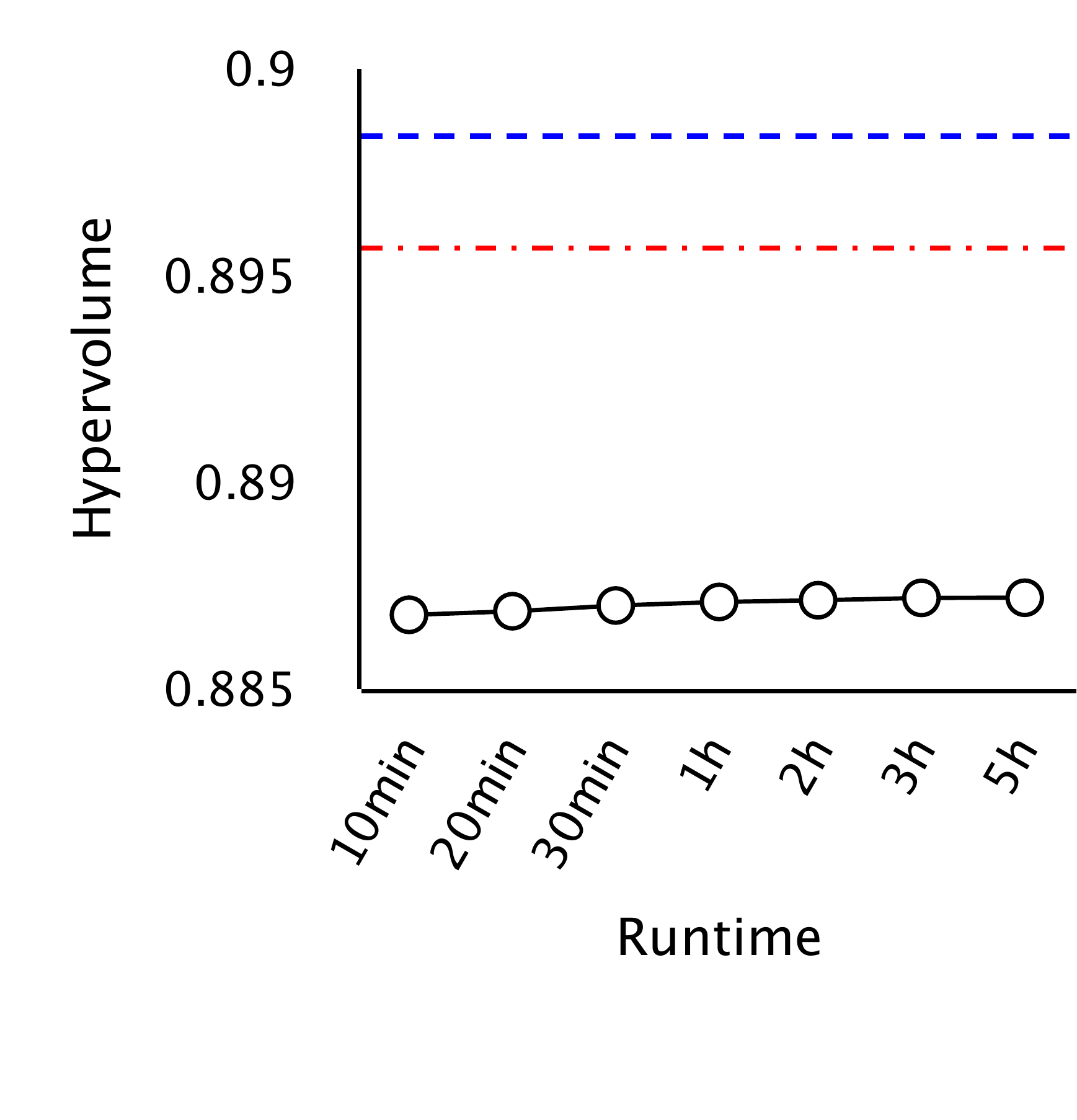}}
    \subfloat[\footnotesize\texttt{a280\_05\_usw\_05}]{\includegraphics[width=0.33\textwidth]{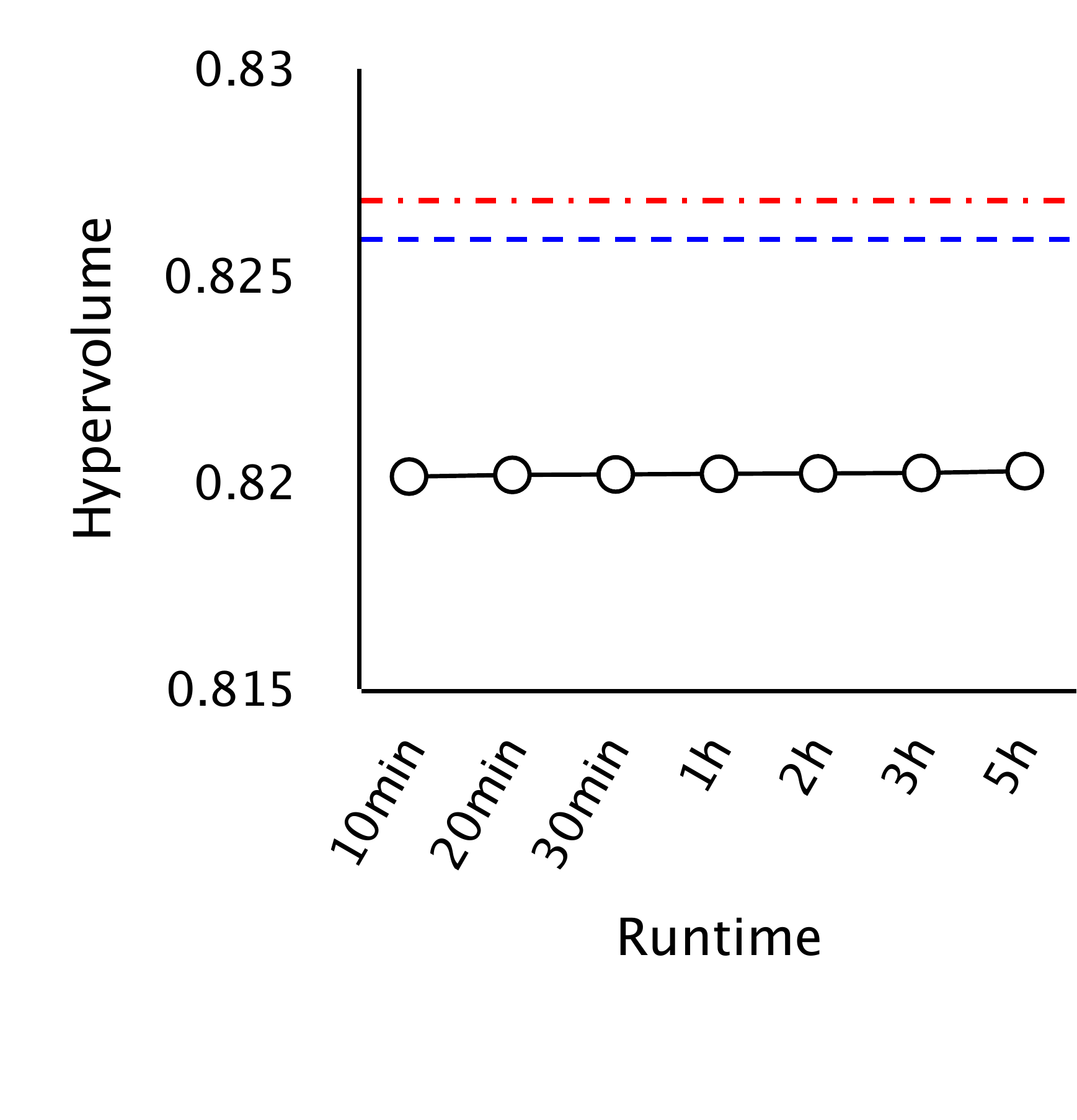}}
    \subfloat[\footnotesize\texttt{a280\_10\_unc\_10}]{\includegraphics[width=0.33\textwidth]{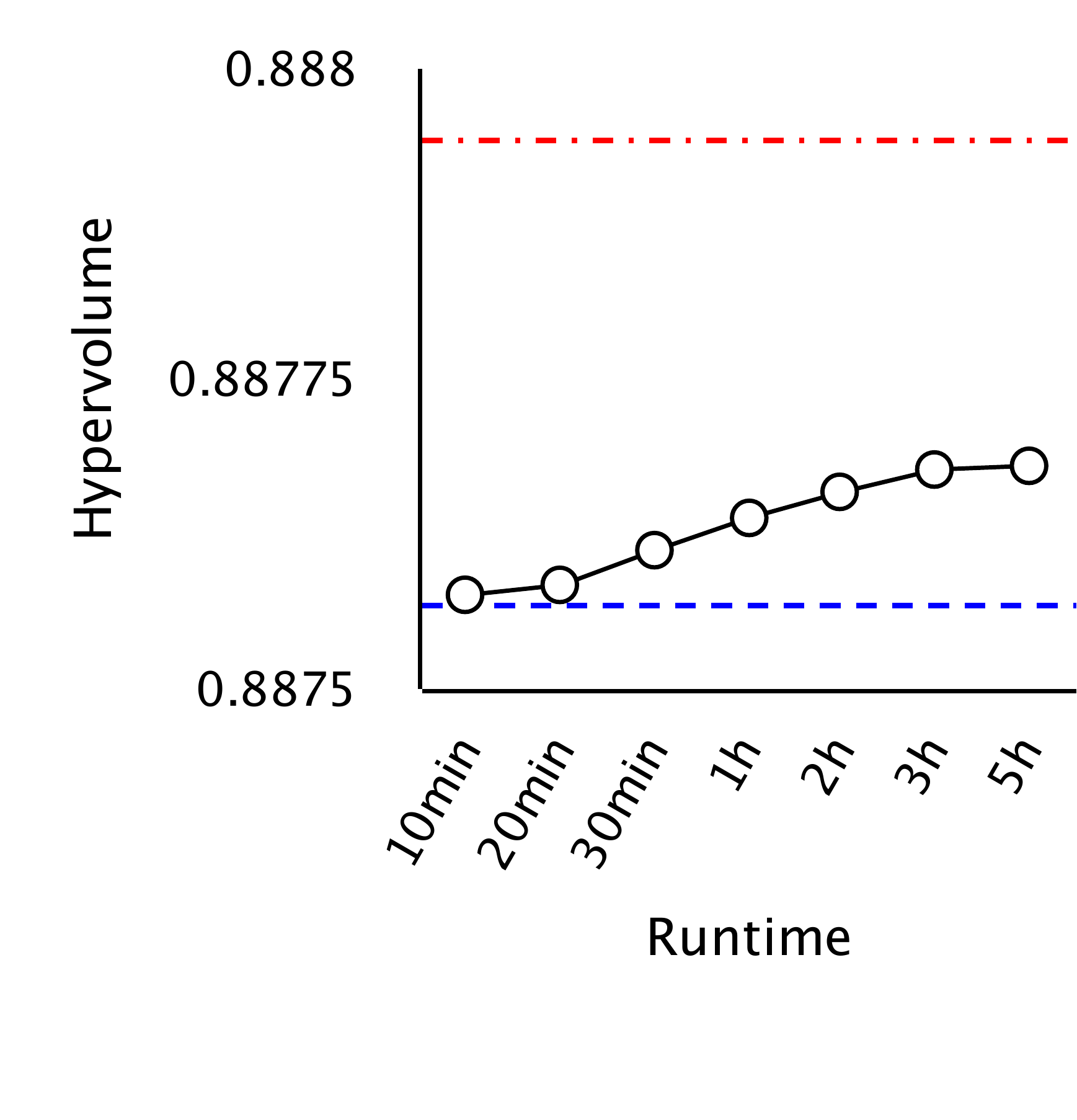}}
    
    \subfloat[\footnotesize\texttt{fnl4461\_01\_bsc\_01}]{\includegraphics[width=0.33\textwidth]{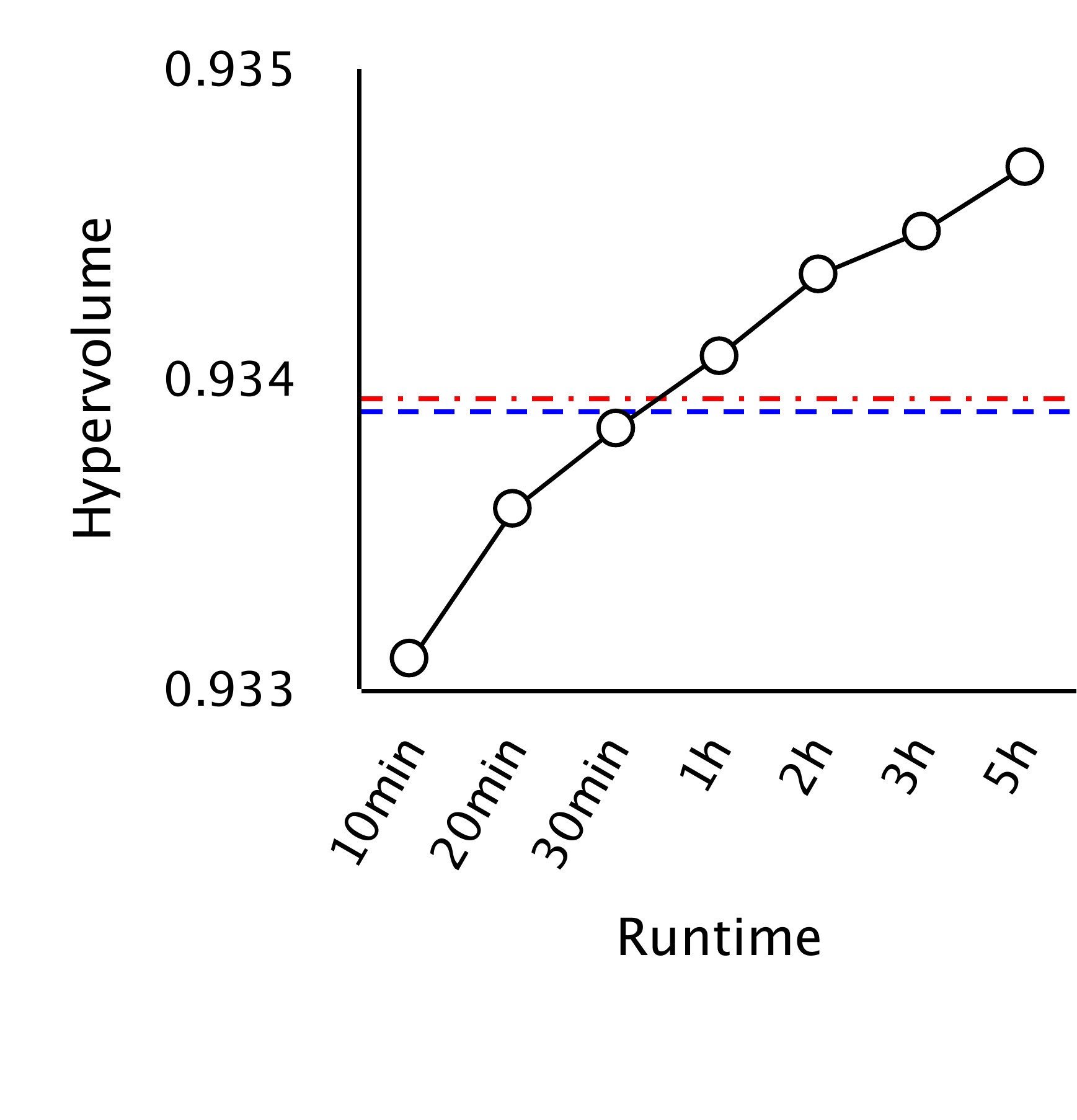}}
    \subfloat[\footnotesize\texttt{fnl4461\_05\_usw\_05}]{\includegraphics[width=0.33\textwidth]{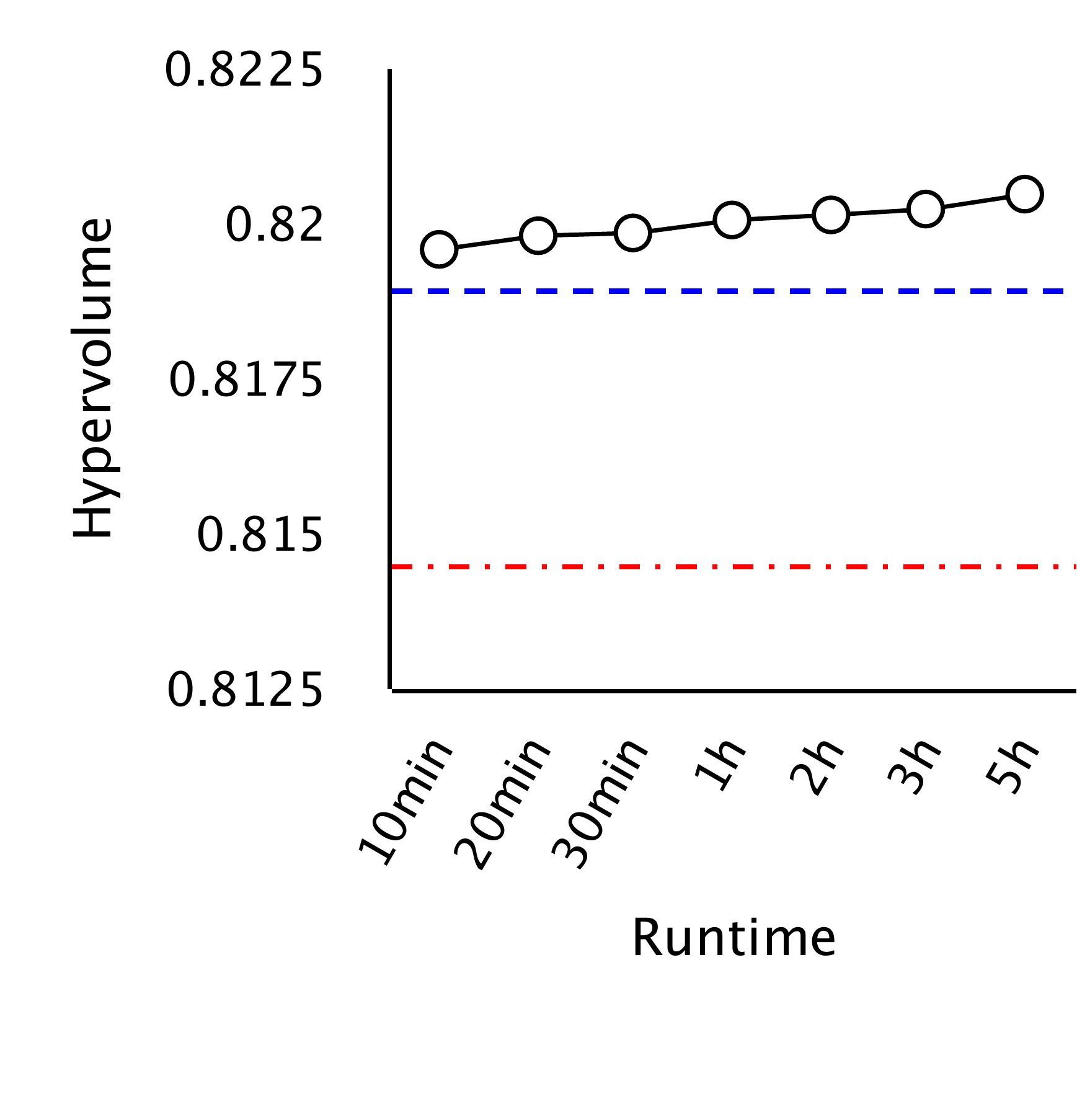}}
    \subfloat[\footnotesize\texttt{fnl4461\_10\_unc\_10}]{\includegraphics[width=0.33\textwidth]{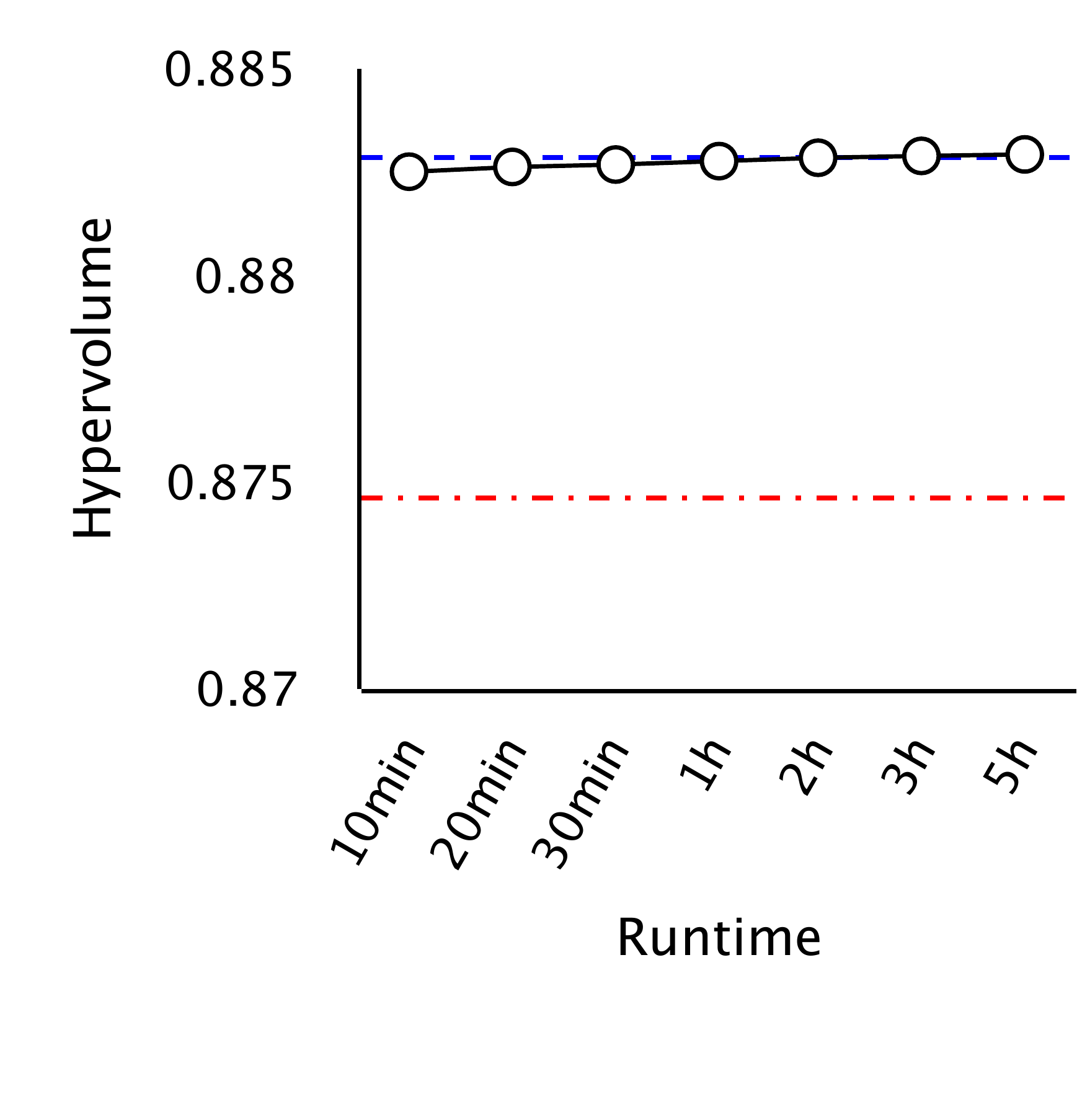}}
    
    \subfloat[\footnotesize\texttt{pla33810\_01\_bsc\_01}]{\includegraphics[width=0.33\textwidth]{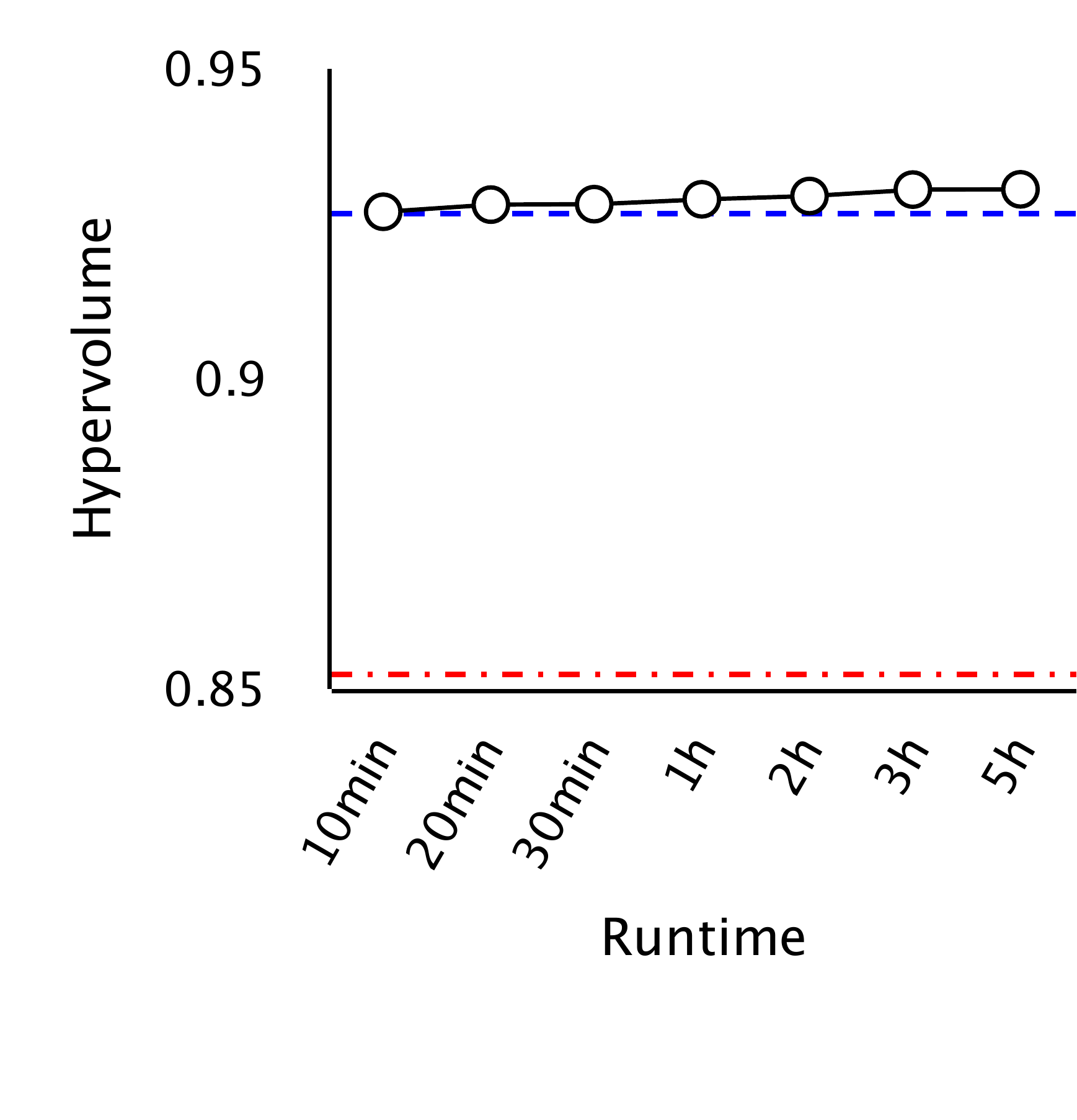}}
    \subfloat[\footnotesize\texttt{pla33810\_05\_usw\_05}]{\includegraphics[width=0.33\textwidth]{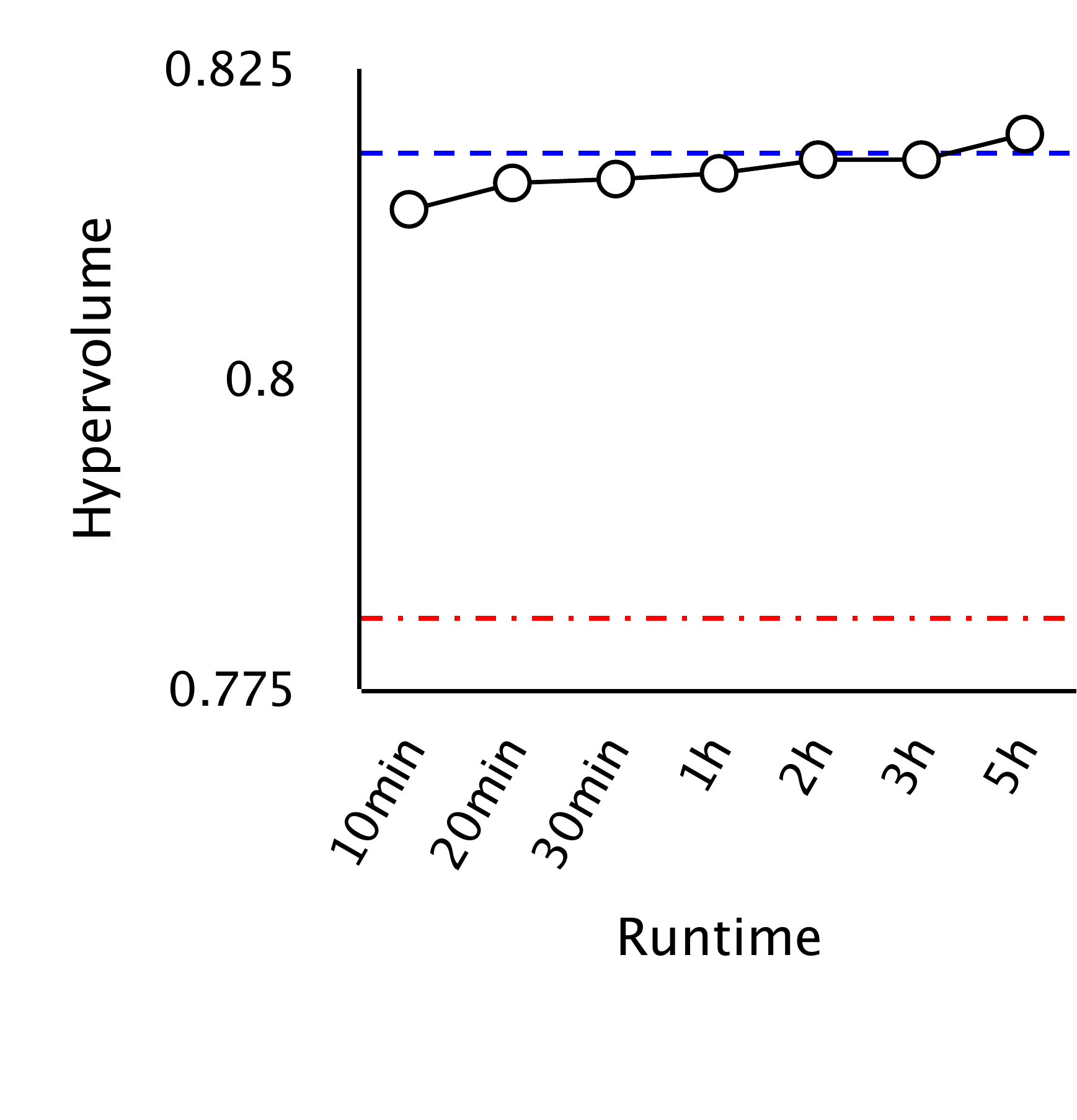}}
    \subfloat[\footnotesize\texttt{pla33810\_10\_unc\_10}]{\includegraphics[width=0.33\textwidth]{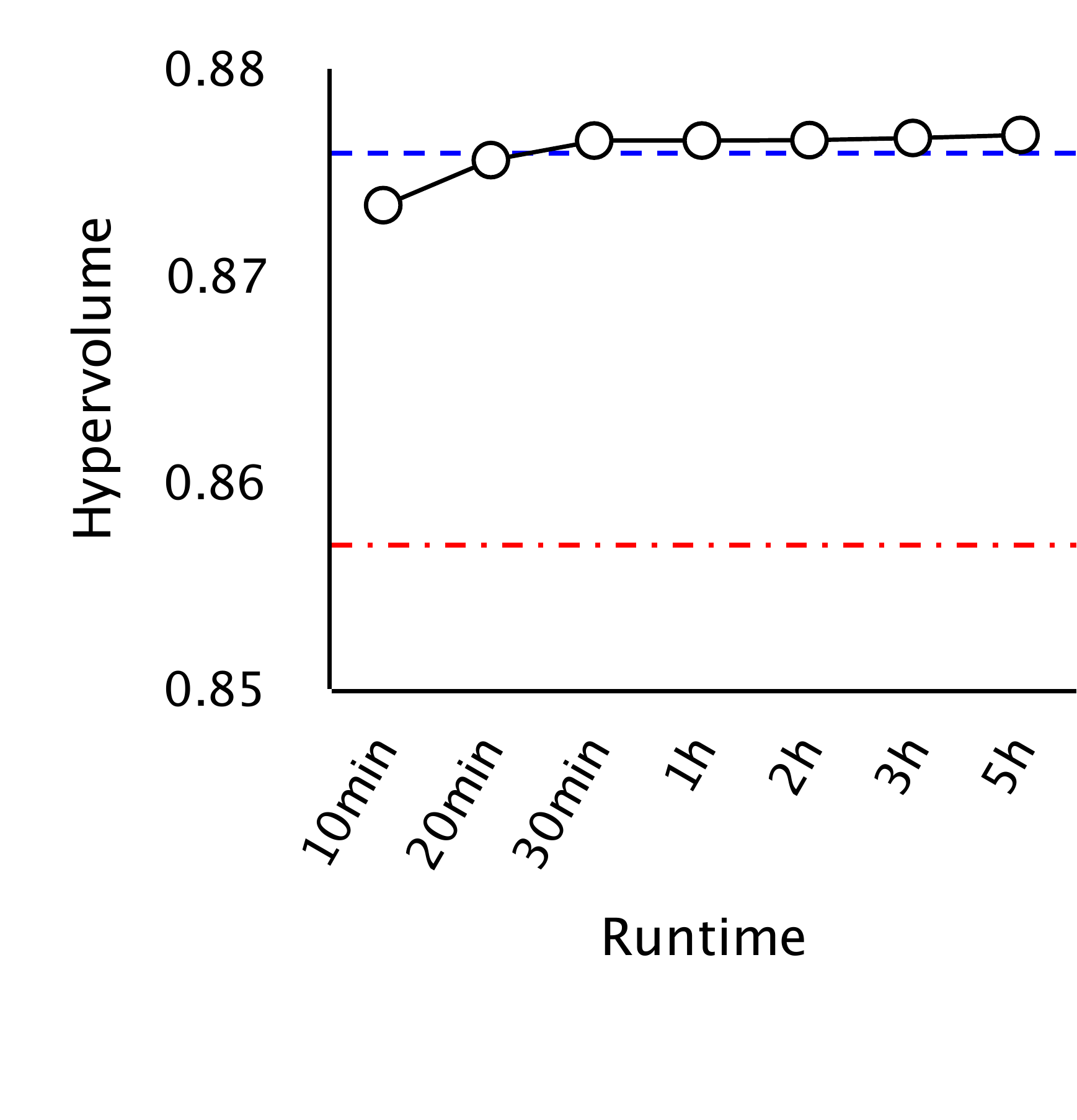}}
    \vspace{-1.0cm}
    \subfloat[]{\includegraphics[width=0.15\textwidth]{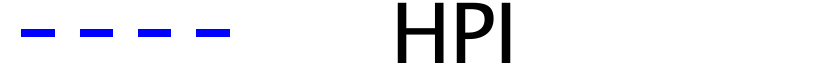}}
    \subfloat[]{\includegraphics[width=0.15\textwidth]{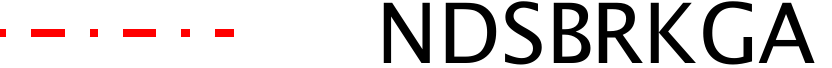}}
    \caption{Hypervolume of WSM over time versus the  hypervolume of HPI and NDSBRKGA; for the latter two, only the final hypervolume is known.}
    \label{fig:wsm_hv_over_time}
\end{figure*}

From Table~\ref{table:competition_results}, we can notice that WSM has obtained better performance on large-size instances. For the three smallest instances, it has presented the worst results concerning the other results, especially, for those reached by the first (HPI) and second (NDSBRKGA) places at GECCO2019 competition. For the other instances, our results have surpassed all other submissions with a larger difference compared to NDSBRKGA. In Figure~\ref{fig:wsm_hv_over_time}, we show the hypervolume achieved by WSM over runtime. In order to make a visual comparison, we have plotted on horizontal lines the (final) hypervolume achieved by the two best algorithms (HPI and NDSBRKGA) of the competitions. One can see that our WSM is able to find solutions that cover a high hypervolume even with low computational time.

\subsection{Dispersed distribution of the non-dominated solutions}

\begin{figure*}%[!ht]
    \captionsetup[subfigure]{position=top, justification=centering, labelformat=empty, oneside, margin={0.8cm,0cm}}
    \centering
    \subfloat[\footnotesize\texttt{a280\_01\_bsc\_01}]{\includegraphics[width=0.33\textwidth]{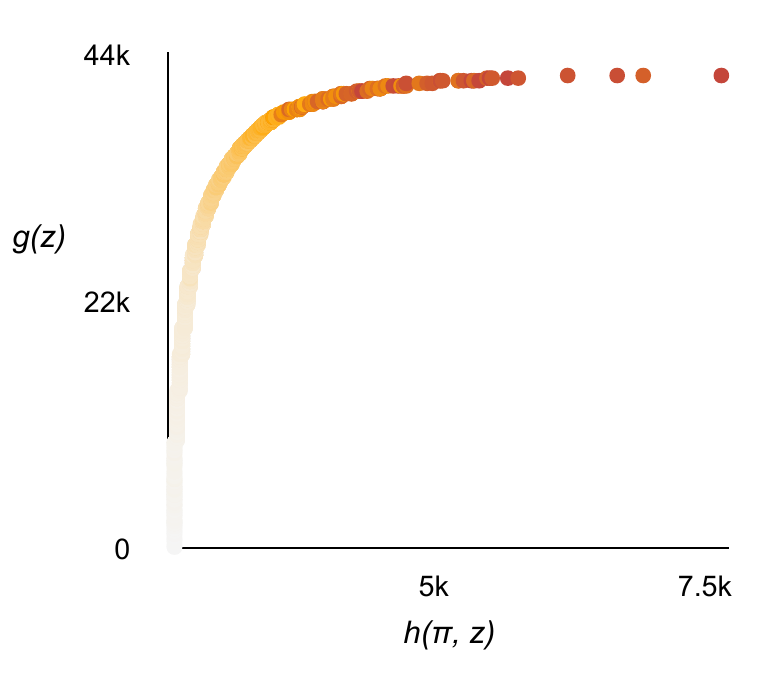}}
    \subfloat[\footnotesize\texttt{a280\_05\_usw\_05}]{\includegraphics[width=0.33\textwidth]{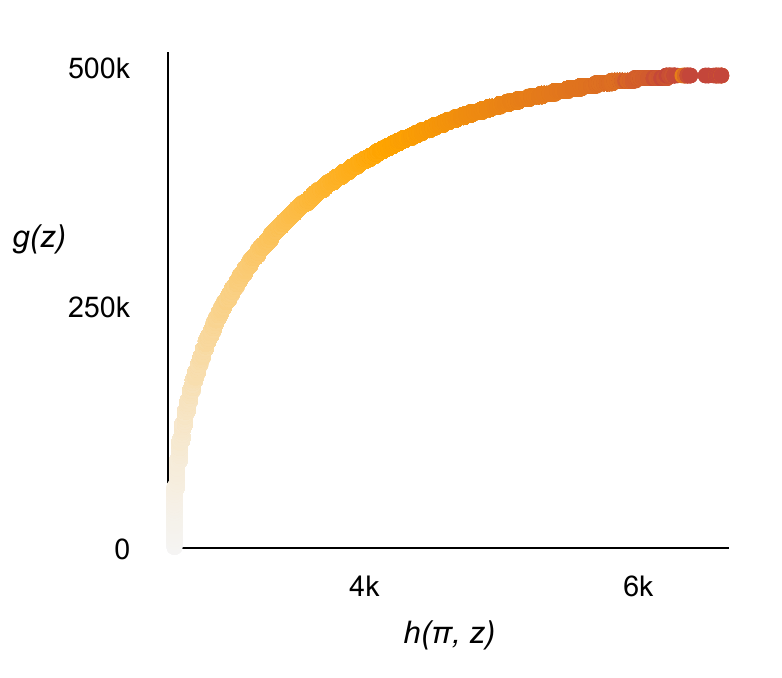}}
    \subfloat[\footnotesize\texttt{a280\_10\_unc\_10}]{\includegraphics[width=0.33\textwidth]{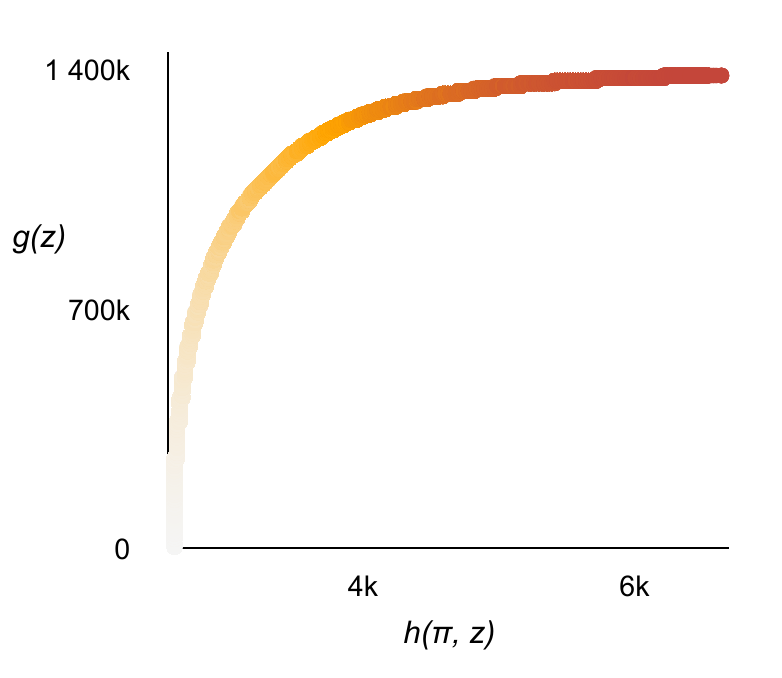}}
    
    \subfloat[\footnotesize\texttt{fnl4461\_01\_bsc\_01}]{\includegraphics[width=0.33\textwidth]{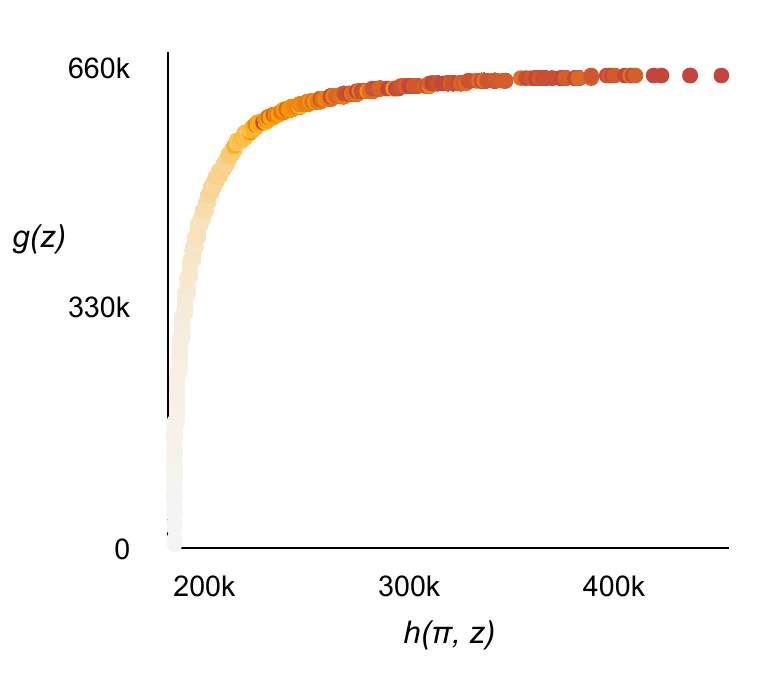}}
    \subfloat[\footnotesize\texttt{fnl4461\_05\_usw\_05}]{\includegraphics[width=0.33\textwidth]{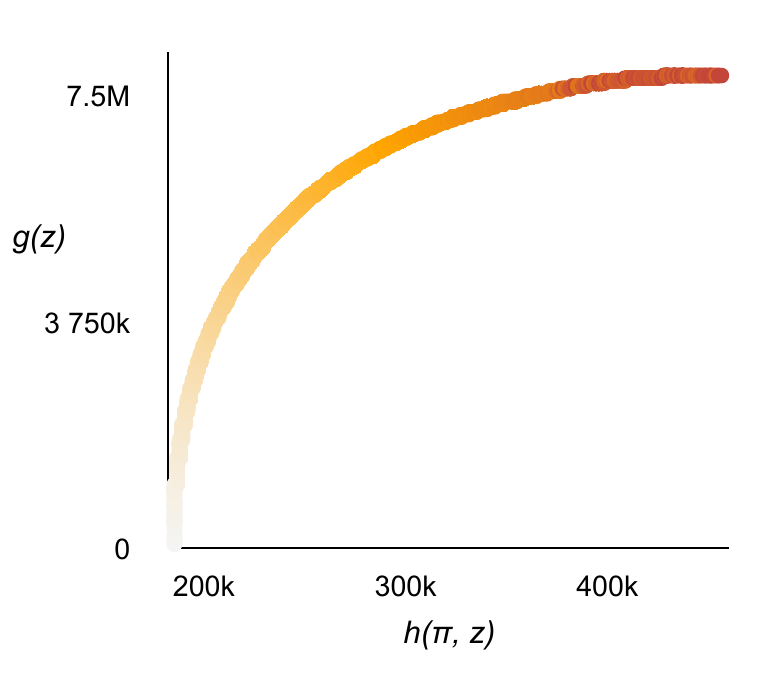}}
    \subfloat[\footnotesize\texttt{fnl4461\_10\_unc\_10}]{\includegraphics[width=0.33\textwidth]{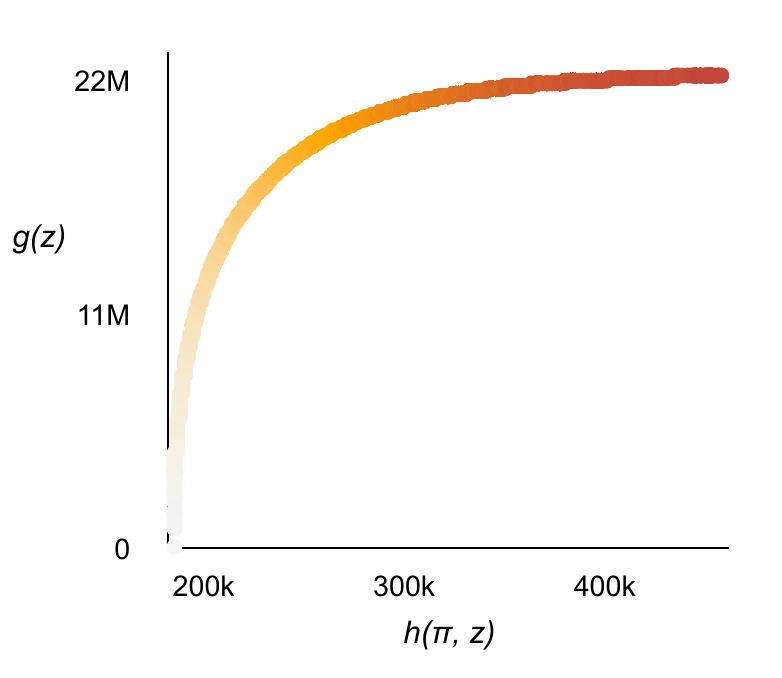}}
    
    \subfloat[\footnotesize\texttt{pla33810\_01\_bsc\_01}]{\includegraphics[width=0.33\textwidth]{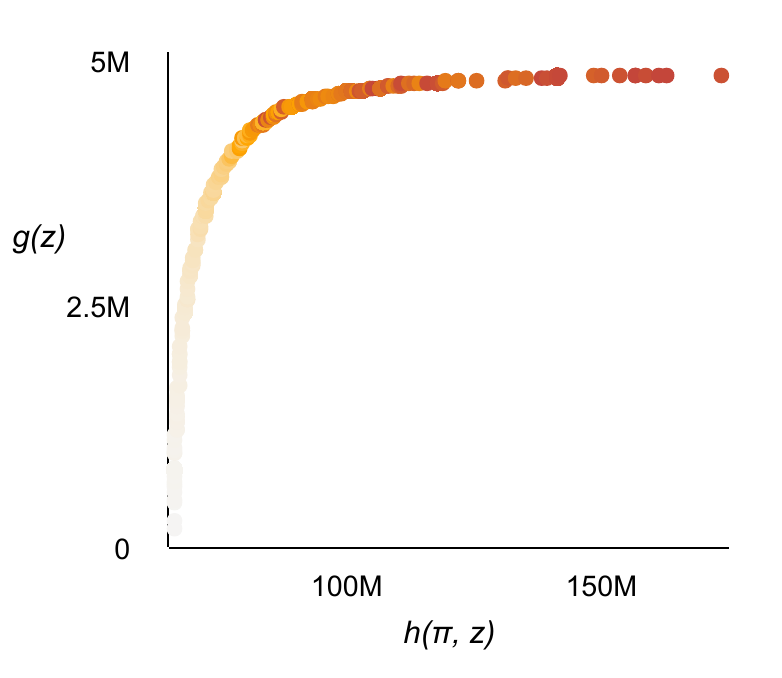}}
    \subfloat[\footnotesize\texttt{pla33810\_05\_usw\_05}]{\includegraphics[width=0.33\textwidth]{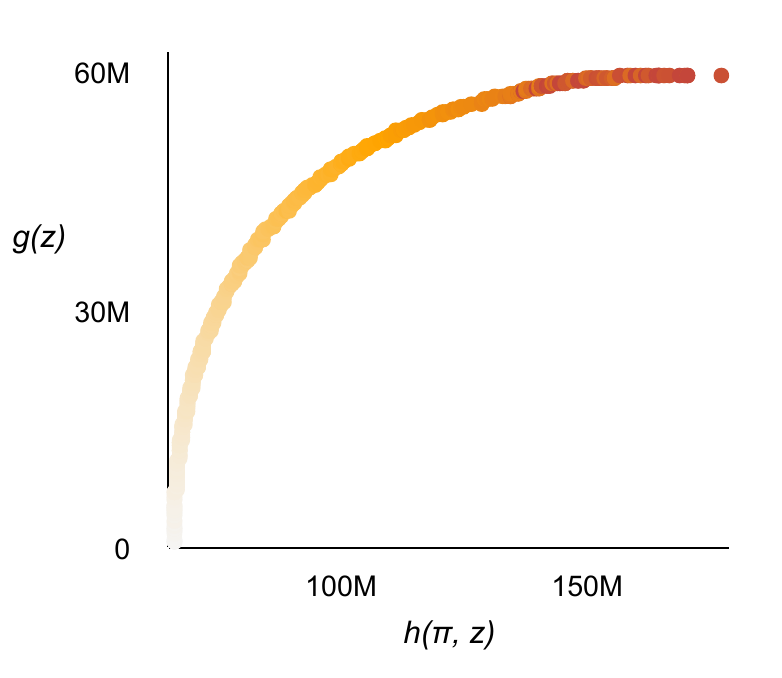}}
    \subfloat[\footnotesize\texttt{pla33810\_10\_unc\_10}]{\includegraphics[width=0.33\textwidth]{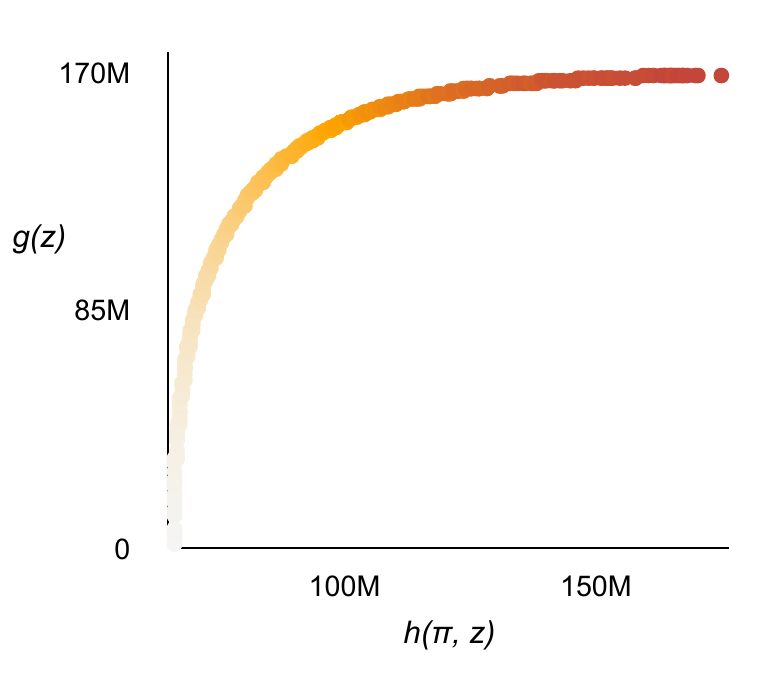}}
    \vspace{-0.5cm}
    \subfloat[]{\includegraphics[width=0.8\textwidth]{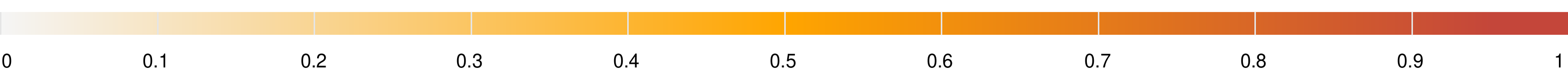}}
    \caption{Non-dominated points found by WSM. Colors indicate the $\alpha$ values used when finding each point.}
    \label{fig:non_dominated_points}
\end{figure*}

We now analyze the dispersion over the objective spaces of the solutions found by our algorithm. As we have stated before, a limitation of WSMs is the fact that, even with a consistent change in weights attributed to the objectives,  they may not generate a dispersed distribution of non-dominated solutions found. This limitation does not affect our WSM, as it can be seen in Figure~\ref{fig:non_dominated_points}, where we have plotted the objective values of all non-dominated solutions found by WSM with 10 minutes of runtime for the nine medium/large-size instances used in the aforementioned BITTP competitions. In addition, we have highlighted which $\alpha$ has been used when finding each solution. One can notice dispersed distributions of the solutions as well as the $\alpha$ values. Moreover, as expected, lower $\alpha$ values produce solutions with faster tours, with higher ones produce solutions with good packing plans.

\subsection{Single-objective comparison}

Since BITTP is a bi-objective formulation created from the TTP without introducing any new specification or removing any original constraint, any feasible BITTP solution is also feasible for the TTP. Thus, we can measure the performance of the solutions obtained by our algorithm according to their single-objective TTP scores. However, it is important to emphasize that our algorithm has not been developed with a single-objective purpose. Therefore, we should be careful when comparing it with other algorithms for the TTP. 

\begin{figure}%[!ht]
    \centering
    \includegraphics[width=0.82\textwidth]{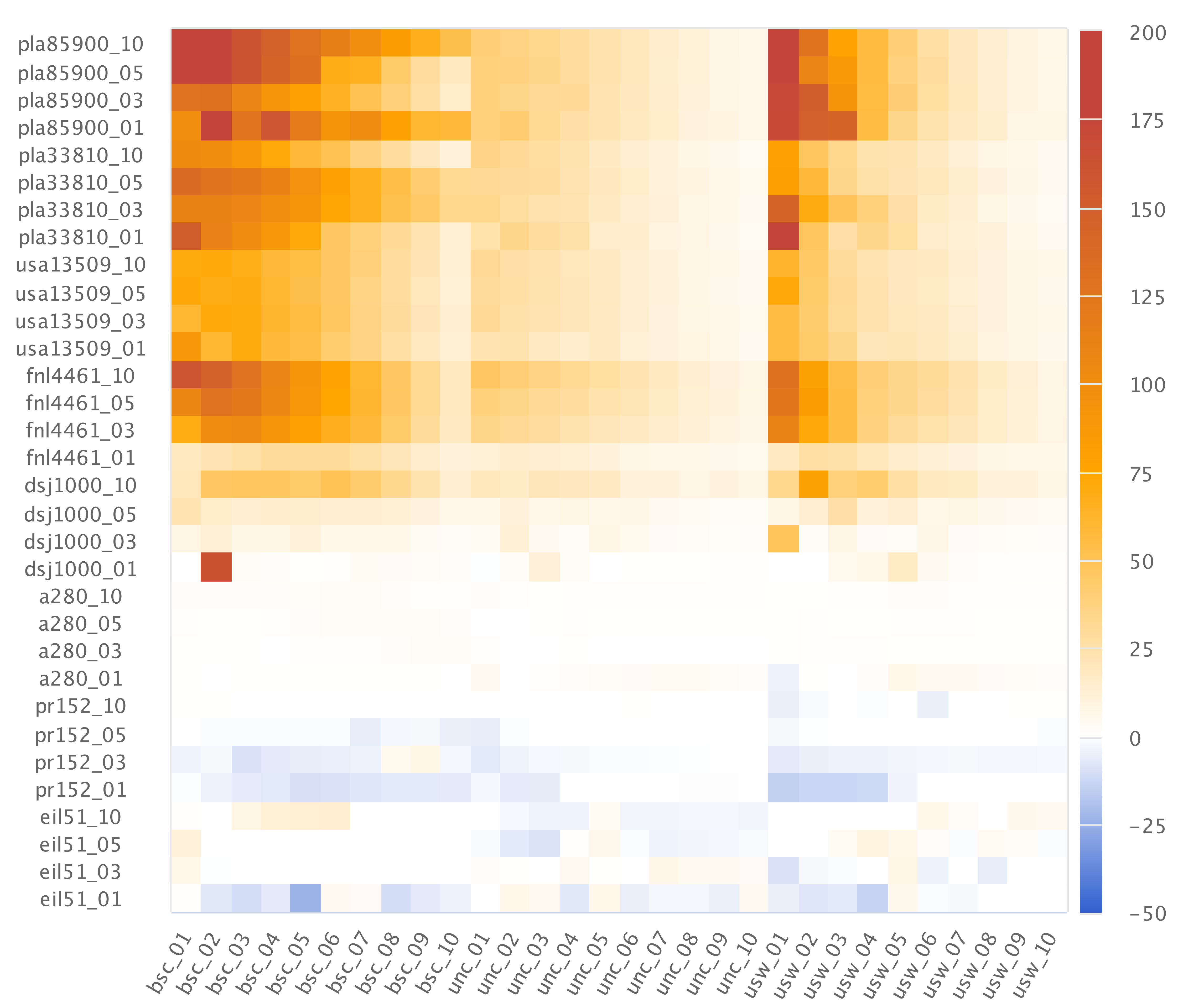}
    \caption{WSM \textit{vs.} NDSBRKGA according to their obtained single-objective TTP scores. Shades of orange and red indicate in which instances our WSM has reached better single-objective TTP scores than NDSBRKGA, while shades of blue indicate the opposite.}
    \label{fig:wsm_vs_ndsbrkga_ttp_scores}
\end{figure}

A fairer comparison can be achieved between our results and those reached by NDSBRKGA, as both approaches have been developed with the same ambition. For this purpose, we have calculated for each instance the Relative Percentage Difference (RPD) between the best TTP scores achieved by WSM and NDSBRKGA, referenced as $\text{S}_{\text{best}}^{\text{WSM}}$ and $\text{S}_{\text{best}}^{\text{NDSBRKGA}}$, respectively. It is important to emphasize that no additional tests have been performed, we only choose the solution with the best TTP score among the non-dominated solutions found by each algorithm on each instance. The RPD metric has been calculated as 
\begin{equation}
    \big(\text{S}_{\text{best}}^{\text{WSM}} - \text{S}_{\text{best}}^{\text{NDSBRKGA}}\big)\;\big/\;\big\rvert \text{S}_{\text{best}}^{\text{NDSBRKGA}}\big\lvert \cdot 100\% \nonumber
\end{equation}

\noindent, and we plot its values using a heatmap in order to highlight higher differences as depicted in Figure~\ref{fig:wsm_vs_ndsbrkga_ttp_scores}. Note that positive values (highlighted in shades of orange and red) indicate that our WSM has found higher TTP scores. 

We can note that the heatmap show in Figure~\ref{fig:wsm_vs_ndsbrkga_ttp_scores} has characteristics similar to those in Figure~\ref{fig:percentage_number_nds}b, where higher percentages of the number of non-dominated solutions found by our algorithm are highlighted. Therefore, this behavior is not surprising, since dominated solutions have essentially lower TTP scores compared to the non-dominated solutions. Thus, we can confirm a better efficiency of WSM also concerning the TTP scores for larger instances, while its worst performance on smaller-size instances is less expressive.

\begin{figure}%[!ht]
    \centering
    \includegraphics[width=0.82\textwidth]{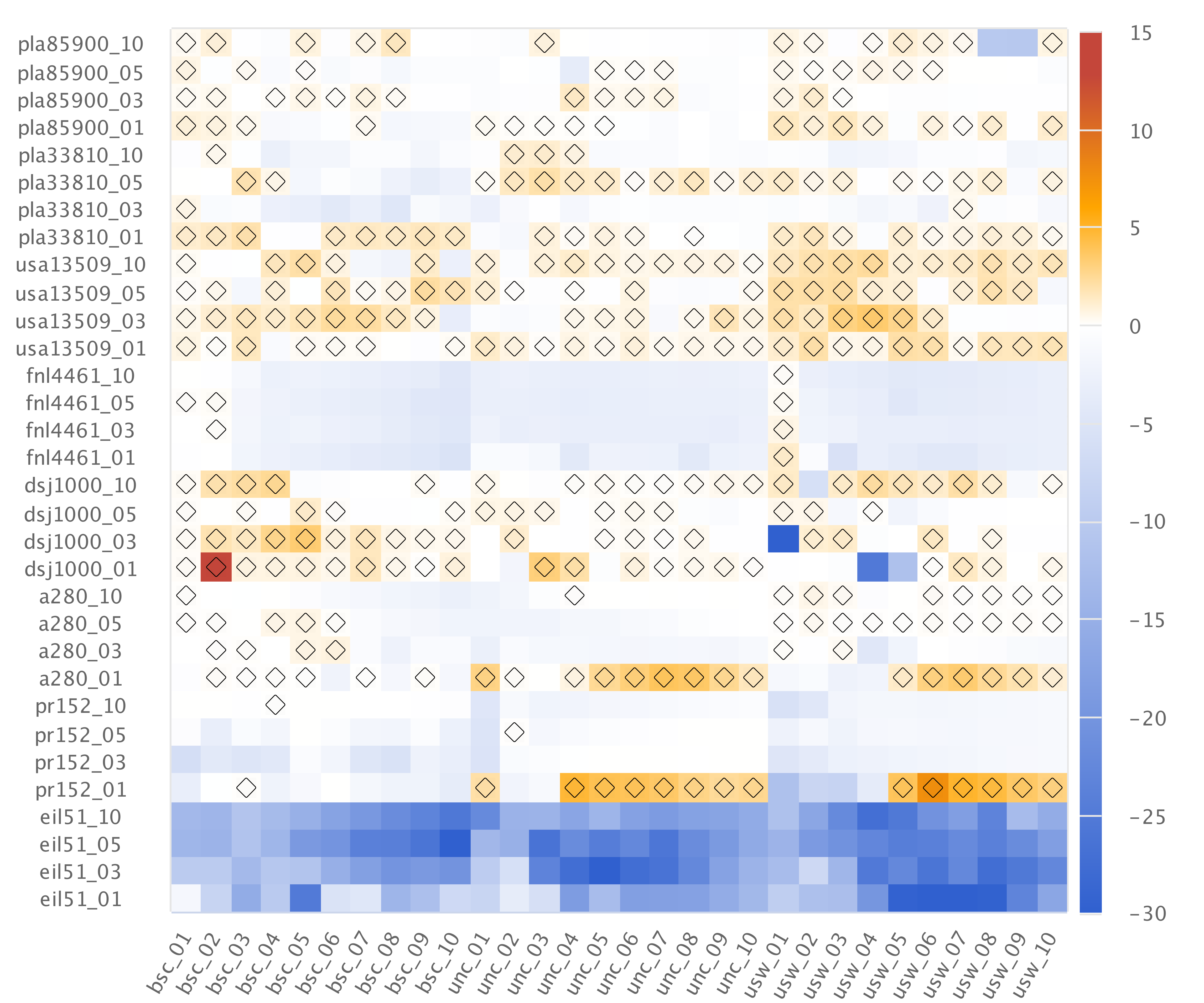}
    \caption{WSM \textit{vs.} TTP algorithms according to their obtained single-objective TTP scores. Shades of orange and red indicate in which instances our WSM has reached better single-objective TTP scores than the best algorithm among 21 ones reported in \citep{wagner2018case}, while shades of blue indicate the opposite. Diamond symbols highlight the 379 instances on which our WSM has found better results.}
    \label{fig:wsm_vs_ttp_scores}
\end{figure}

Although it may not be fair, as we stated earlier, we conclude our analysis by comparing the best TTP scores obtained by WSM with the best single TTP objective scores reported in \citep{wagner2018case}, where the authors have made a comprehensive comparison of 21 algorithms proposed for the TTP over the years. In this comparison, we use again the RPD metric, which now is calculated for each instance as
\begin{equation}
    \big(\text{S}_{\text{best}}^{\text{WSM}} - \text{S}_{\text{best}}^{\text{21ALGS}}\big)\;\big/\;\big\rvert \text{S}_{\text{best}}^{\text{21ALGS}}\big\lvert \cdot 100\% \nonumber
\end{equation}

\noindent, where $\text{S}_{\text{best}}^{\text{21ALGS}}$ indicates the best TTP score found among all 21 algorithms analyzed in \citep{wagner2018case}. In Figure~\ref{fig:wsm_vs_ttp_scores}, we plot the calculated RPD values following the same visualization scheme adopted previously. In addition, we highlight with a diamond symbol the instances for which our algorithm has found better solutions.

One can note that, in general, our results presented worse performance, which is especially true for the smaller-size instances. However, for 379 instances our results have outperformed all 21 TTP algorithms. This shows that our WSM can also be competitive to solve the TTP.

\section{Conclusions}\label{sec:conclusions}

In this work, we have addressed a bi-objective formulation of the Traveling Thief Problem (TTP), an academic multi-component problem that combines two classic combinatorial optimization problems: the Traveling Salesperson Problem and the Knapsack Problem. 
For solving the problem, we have proposed a heuristic algorithm based on the well-known weighted-sum method, in which the objective functions are summed up with varying weights and then the problem is optimized in relation to the single-objective function formed by this sum. 
Our algorithm combines exploration and exploitation search procedures by using efficient operators, as well as known strategies for the single-objective TTP; among these are deterministic strategies that we have randomized here. 
We have studied the effects of our algorithmic components by performing extensive tuning of their parameters over different groups of instances. 
This tuning also shows that different configurations are needed depending on the instance group, the knapsack type, and the knapsack capacity. 
Our comparison with multi-objective approaches shows that we outperform participants of recent optimization competitions, and we have furthermore found new best solutions for 379 instances to the single-objective case along the way.

For future research, we would like to point out as a promising direction the investigation of the influence of different algorithmic components already proposed in the literature over different instance characteristics by investigating tuned configurations. Studies in this data-driven direction have achieved important insights to design better single-objective solvers for fundamental problems and real-world problems (see, e.g. Section ``Research Directions'' of \cite{Agrawal2020duo}). Another interesting direction would be to use our algorithm core idea for solving other multi-objective problems with multiple interacting components. By core idea, we refer to how to explore and exploit the space of solutions once efficient operators and strategies are known for solving different components of a multi-objective problem.

\vspace{2mm}\noindent\textbf{Acknowledgments.}
This study has been financed in part by Coordena\c{c}\~{a}o de A\-per\-fei\-\c{c}o\-a\-men\-to de Pessoal de N\'{i}vel Superior - Brazil (CAPES) - Finance code 001. The authors would also like to thank Funda\c{c}\~{a}o de Amparo \`{a} Pesquisa do Estado de Minas Gerais (FAPEMIG), Conselho Nacional de Desenvolvimento Cient\'{i}fico e Tecnol\'{o}gico (CNPq), Universidade Federal de Ouro Preto (UFOP) and Universidade Federal de Vi\c{c}osa (UFV) for supporting this research. Markus Wagner would like to acknowledge support by the Australian Research Council Project DP200102364.

%\clearpage


%\FloatBarrier
% \bibliographystyle{elsarticle-num}
% \bibliographystyle{elsarticle-harv}
% \bibliographystyle{elsarticle-num-names}\biboptions{authoryear}
% \bibliographystyle{model1a-num-names}\biboptions{authoryear}
% \bibliographystyle{model1b-num-names}\biboptions{authoryear}
% \bibliographystyle{model1c-num-names}\biboptions{authoryear}
% \bibliographystyle{model1-num-names}\biboptions{authoryear}
\bibliographystyle{model2-names}\biboptions{authoryear}

\bibliography{bibfile}

\end{document}